\documentclass[letterpaper]{article} 
\usepackage{aaai24}  
\usepackage{times}  
\usepackage{helvet}  
\usepackage{courier}  
\usepackage[hyphens]{url}  
\usepackage{graphicx} 
\usepackage{natbib}  
\usepackage{caption} 
\usepackage{algorithm}
\usepackage{algorithmic}
\usepackage{newfloat}
\usepackage{listings}
\DeclareCaptionStyle{ruled}{labelfont=normalfont,labelsep=colon,strut=off} 
\usepackage{color}
\usepackage{colortbl}
\usepackage{xcolor}
\usepackage{comment}
\usepackage{enumerate}
\usepackage{multirow}
\usepackage{multicol}
\usepackage{indentfirst}
\usepackage{booktabs}
\usepackage{subfigure}
\usepackage{tikz}
\usepackage{cancel}
\usepackage{gensymb}
\usepackage{dsfont}
\usepackage{mathtools}
\usepackage{caption}
\usepackage{textcomp}
\usepackage{longtable}
\usepackage{natbib} 
\usepackage{caption}
\usepackage{epsfig}
\usepackage{bbding}
\usepackage{pifont}
\usepackage{fontawesome}
\usepackage{wasysym}
\usepackage[mathscr]{eucal}
\usepackage{amssymb} 
\usepackage{arydshln}
\usepackage{makecell}
\usepackage[utf8]{inputenc}
\usepackage{natbib}  
\usepackage[super]{nth}
\newcommand{\tabincell}[2]{\begin{tabular}{@{}#1@{}}#2\end{tabular}}
\usepackage{pifont}
\definecolor{mygreen}{RGB}{81,150,111}
\definecolor{myred}{RGB}{160,0,0}
\definecolor{myblue}{RGB}{89,158,254}
\definecolor{myblue}{RGB}{89,158,254}

\usepackage[capitalize]{cleveref}
\crefname{section}{Sec.}{Secs.}
\Crefname{section}{Section}{Sections}
\Crefname{table}{Table}{Tables}
\crefname{table}{Tab.}{Tabs.}

\floatstyle{ruled}
\newfloat{listing}{tb}{lst}{}
\floatname{listing}{Listing}
\title{FastPillars: A Deployment-friendly Pillar-based 3D Detector}
\author{
    Sifan Zhou\textsuperscript{\rm 1}, 
    Zhi Tian\textsuperscript{\rm 2}, 
    Xiangxiang Chu\textsuperscript{\rm 2}, 
    Xinyu Zhang\textsuperscript{\rm 2}, 
    Bo Zhang\textsuperscript{\rm 2}, 
    Xiaobo Lu$^{1}$\footnotemark[1]\\
    Chengjian Feng\textsuperscript{\rm 2},  
    Zequn Jie\textsuperscript{\rm 2},  
    Miao Sun\textsuperscript{\rm 3},  
    Patrick Yin Chiang\textsuperscript{\rm 3},
    Lin Ma\textsuperscript{\rm 2}
}
\affiliations{
    \textsuperscript{\rm 1}Southeast University
    \textsuperscript{\rm 2}Meituan Inc.
    \textsuperscript{\rm 3}Fudan University\\
sifanjay@gmail.com, \{tianzhi02,chuxiangxiang,zhangxinyu35,zhangbo97\}@meituan.com \\
 xblu2013@126.com, \{fengchenhgjian,jiezequn,malin11\}@meituan.com, pchiang@photonic-tech.com}

\usepackage{bibentry}

\begin{document}
\maketitle

\begin{abstract}
The deployment of 3D detectors strikes one of the major challenges in real-world self-driving scenarios. Existing BEV-based (i.e., Bird Eye View) detectors favor sparse convolutions (known as SPConv) to speed up training and inference, which puts a hard barrier for deployment, especially for on-device applications. In this paper, in order to tackle the challenge of efficient 3D object detection from an industry perspective, we devise a deployment-friendly pillar-based 3D detector, termed FastPillars. First, we introduce a novel lightweight Max-and-Attention Pillar Encoding (MAPE) module specially for enhancing small 3D objects. Second, we propose a simple yet effective principle for designing backbone in pillar-based 3D detection. We construct FastPillars based on these designs, achieving high performance and low latency without SPConv. Extensive experiments on two large-scale datasets demonstrate the effectiveness and efficiency of FastPillars for on-device 3D detection regarding both performance and speed. Specifically, FastPillars delivers state-of-the-art accuracy on Waymo Open Dataset with 1.8 $\times$ speed up and 3.8 mAPH/L2 improvement over CenterPoint (SPConv-based). We will release our code.
\end{abstract}

\section{Introduction}
\label{sec:intro}
3D object detection using LiDAR point cloud has a wide range of applications and has shown remarkable progress in self-driving and robotics~\cite{qi2018frustum, shi2020pv, shi2020points}. However, the community tends to explore high-performance detectors while overlooking the requirement for fast runtime speed, essential for onboard deployment in autonomous systems. Therefore, it is imperative to develop a real-time top-performing 3D detector from the industrial perspective.

According to the type of input data, existing real-time mainstream 3D detectors can be divided into two classes: point-based~\cite{qi2019deep, shi2019pointrcnn,shi2020points, yang20203dssd} and grid-based methods~\cite{zhou2018voxelnet,  lang2019pointpillars, deng2021voxel}. In point-based methods, PointNet families~\cite{qi2017pointnet, qi2017pointnet++} are leveraged to learn discriminative representation from raw point cloud. However, they may not be friendly to effective hardware implementation as they often require point query/retrieval in 3D space (\textit{e.g.}, PointNet++~\cite{qi2017pointnet++}). Voxel-based 3D detectors convert irregular point cloud into arranged grids (\textit{i.e.}, voxels/pillars). However, sparse point cloud result in numerous empty grids, which will lead to significant redundant computational overheads. Some methods~\cite{yan2018second,  yin2021center} use SPConv~\cite{graham2017submanifold} to skip calculation on empty grids to reduce computational burden. Despite being effective, SPConv poses a challenge when converted to TRT (i.e., TensorRT) for deployment and hampers further speedup through these techniques. 

\begin{figure}[t]
  \centering
  \includegraphics[width=\linewidth]{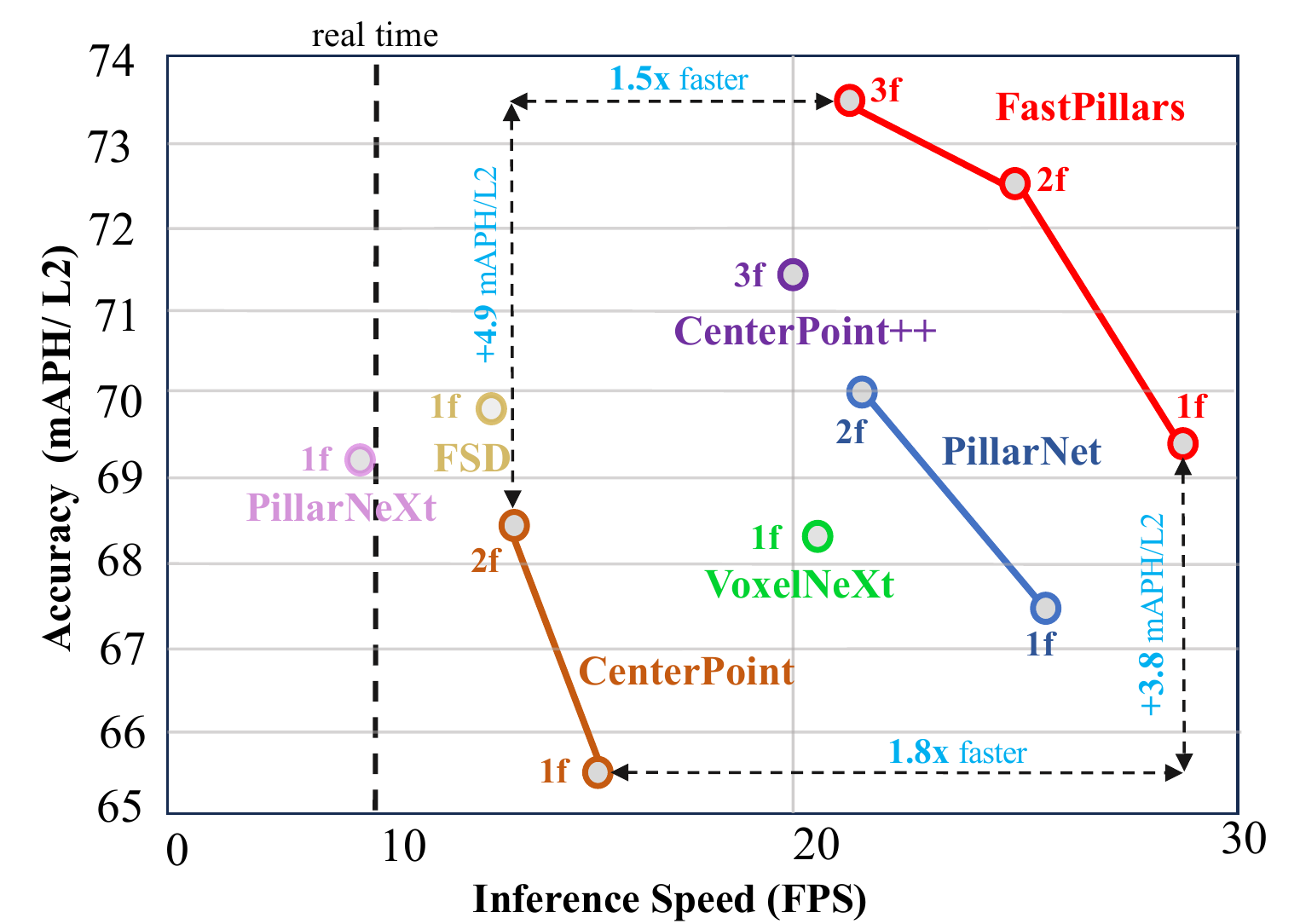}
   \caption{The overall comparison with other real-time one-stage 3D detection methods. FastPillars finds a better trade-off between accuracy and speed, outperforming CenterPoint by 1.8 $\times$ faster and 3.8 mAPH L2 higher on the Waymo \texttt{val} set. FPS is measured on an NVIDIA Tesla V100 GPU.}
\label{fig:compare}
 \vspace{-3.0mm}
\end{figure}
PointPillars~\cite{lang2019pointpillars} is proposed to utilize highly-optimized 2D convolutions alone, achieving lower latency. Despite being fast, PointPillars has unsatisfactory performance due to the lack of an efficient backbone. Recently, PillarNet~\cite{shi2022pillarnet} achieved high-performance while keeping real-time speed based on a SPConv-based backbone. Nevertheless, the use of SPConv in PillarNet makes it hard to be quantized and deployed via TRT. Specifically, the deployment of SPConv has the following difficulties. \textbf{\romannumeral1)} SPConv is not a built-in operation in TensorRT. This makes it necessary to write a tedious custom plugin in CUDA C++ with several limitations like fixed-shape input and reduced compatibility for commonly-used TensorRT for the quantization deployment. \textbf{\romannumeral2)} SPConv requires irregular memory access patterns, which is challenging to optimize on modern hardware. \textbf{\romannumeral3)} SPConv takes as inputs 3D coordinates that are difficult to be quantized and requires taking fusion into consideration for fast execution, further complicating the deployment process. However, TRT offers several advantages: \textbf{\romannumeral1)} TRT can significantly accelerate model inference while maintaining accuracy through deep optimization and high parallelism techniques such as layer and tensor fusion. \textbf{\romannumeral2)} TRT supports a wide range of embedded and automotive environments, making it extensively used in the deployment of deep learning models in various industrial products.

\begin{figure*}[!htb]
  \centering
  \includegraphics[width=0.9\textwidth]{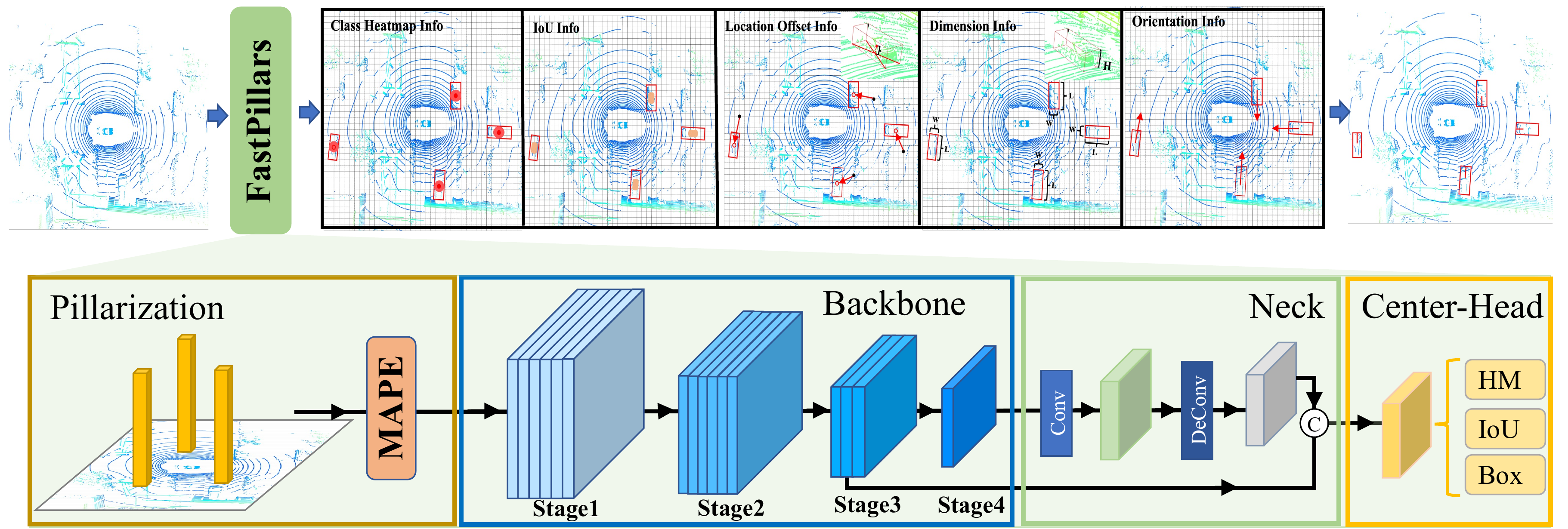}
  \vspace{-1.0mm}
  \caption{The framework of \textbf{FastPillars}. Input the point cloud, \textbf{FastPillars} predicts 3D bounding boxes. As shown in the bottom, \textbf{FastPillars} consists of four parts: \textbf{MAPE} module, backbone, neck and center-based head. First, the point cloud is pillarized with MAPE, and then the encoded features are sent to the backbone for choreographed feature extraction. These features are fused by the neck, and 3D boxes are regressed based on the center-based head. The backbone is designed to be scalable by changing the number of blocks at the early stages according to different needs. Best viewed in color.}
  \label{fig:method}
  \vspace{-1em}
\end{figure*}

In this paper, we propose a real-time and high-performance 3D object detector designed with deployment in mind, termed \textbf{FastPillars}. FastPillars is fully based on standard convolutions, and thus it can be effortlessly deployed in onboard applications and seamlessly enjoy the speedup of TRT and network quantization. FastPillars consists of four essential blocks which are PFE (Pillar Feature Encoding), backbone, neck and head (see Fig. \ref{fig:method}), respectively. In PFE block, we observe that previous pillar-based methods did not pay attention to local geometry patterns. To this end, we propose a simple but effective \textbf{M}ax-and-\textbf{A}ttention \textbf{P}illar \textbf{E}ncoding (MAPE) module, which attentively integrates significant local features and thus alleviates the information loss in the pillar generating process. In addition, the MAPE module barely increases the overall latency (+4ms). In backbone, we find that the design rules of 2D backbone are not suitable for point cloud detection task. Based on this observation, we propose a computation-efficient principle and design a lightweight backbone. Finally, we enrich the semantic features in neck block and adopt center-based detection head. Extensive experiments demonstrate that FastPillars achieves state-of-the-art performance on two large-scale datasets nuScenes and Waymo. As shown in Fig. \ref{fig:compare}, it can be seamlessly speeded up through TensorRT with 1.8 $\times$ speed up and 3.8 mAPH L2 improvement over CenterPoint (SPConv-based). As a result, the proposed method offers an even better trade-off between speed and accuracy for real-time embedded applications. We summarize our contributions as follows:
\begin{itemize}
    \item We introduce a novel lightweight Max-and-Attention Pillar Encoding (\textbf{MAPE}) module specially for enhancing small 3D objects.
    \item We propose a simple yet effective principle for designing pillar-based 3D backbone with remarkable performance and real-time inference speed.
    \item Based on the above designs, we construct a high-performance and low-latency 3D detector termed \textbf{FastPillars}. FastPillars eliminates the need for SPConv hampering the on-device deployment and provides a strong and simple alternative to SPConv-based detectors.
    \item Extensive experiments on the nuScenes and Waymo dataset show that FastPillars brings a new state-of-the-art for on-board 3D detection in terms of accuracy and latency trade-off, and most importantly \emph{end-to-end deployable in TensorRT}. We believe our method serve as a strong competitor to its peers and pose a significant impact for the community. 
\end{itemize}

\vspace{-2.0mm}
\section{Related Work}
\textbf{Voxel-based 3D Detectors.}
Voxel-based 3D detectors~\cite{chen2017multi, kuang2020voxel, deng2021voxel} generally convert the unstructured point cloud to regular pillar/voxel grids. This further allows learning point features by utilizing the mature 2D/3D CNNs. VoxelNet~\cite{zhou2018voxelnet} is a pioneering work, which voxelizes the point cloud and then uses Voxel Feature Extractor (VFE) and 3D CNNs to learn the geometrical representation. Its shortcoming is the slow inference speed due to the huge computational burden of the 3D convolutions. To save the memory cost, SECOND~\cite{yan2018second} uses 3D sparse convolutions~\cite{graham2017submanifold} that operates on non-empty voxels to speed up the training and inference. However, the use of SPConv has a drawback: it is not deployment-friendly, makeing it tricky to apply them on embedded systems. To this end, PointPillars~\cite{lang2019pointpillars} was proposed for on-device deployment. In PointPillars, a mature 2D detector pipeline is applied to predict 3D objects, making it easy to be converted into ONNX/TensorRT for deployment. Meanwhile, PointPillars' deployment-friendly nature has made it a popular method in practice. After that, CenterPoint~\cite{yin2021center} was proposed, which uses a nearly real-time and anchor-free pipeline, achieving state-of-the-art performance. Recently, PillarNet~\cite{shi2022pillarnet} uses 2D SPConv based on the ``encoder-neck-head" architecture to boost accuracy with real-time speed. However, due to the use of SPConv, it inevitably faces the difficulty of deployment for industrial applications and further speedup with TRT optimization.

\textbf{Industry-level Lightweight Network Structures for Object Detection.}
For years, the YOLO series~\cite{bochkovskiy2020yolov4,jocher2022yolov5} has been the \emph{de facto} industry standard for lightweight 2D object detection, whose backbone designs mainly inherit the ideas from RepVGG~\cite{ding2021repvgg}. RepVGG refactored the famous plain network VGG~\cite{simonyan2014very} using a reparameterization-based structural design. During training, a plain Conv-BN-ReLU is replaced by its over-parameterized three-branch counterpart, \textit{i.e.}, Conv3$\times3$-BN, Conv1$\times1$-BN and Identity-BN, followed by the ReLU function after the summation of the three branches. The three-branch structure substantially helps the network optimization while the reparameterization converts three branches identically into one at inference, improving the efficiency in inference. Due to its advantage, this trend has swept 2D object detectors and shown high performance at extreme speeds, such as PPYOLO-E~\cite{xu2022pp}, YOLOv6~\cite{li2022yolov6} and YOLOv7~\cite{wang2022yolov7}. Albeit the success, it is not yet seen, to our best knowledge, any application of these schemes in LiDAR object detection. Our successful application significantly improves computational efficiency and reduces the difficulty of deployment, especially for resource-constrained hardwares.

\vspace{-1.0mm}
\section{Our Approach}
\label{sec:FastPillars}
This section presents FastPillars, an end-to-end trainable and SPConv-free neural network for real-time high-performance 3D detection. As shown in Fig. \ref{fig:method}, our network consists of four blocks: MAPE module, backbone, neck and center-based detection head. 

\textbf{Problem Setting.} We present the basic task definitions of LiDAR-based 3D detection before introducing the detailed method. Given a point set with $N$ points in the 3D space, which is defined as $\mathbf{P}=\{\mathbf{p}_i=[x_i, y_i, z_i, r_i, t_i]^T \in {\mathbb{R}^{N \times 5}} \}$, where $x_i, y_i, z_i$ denote the coordinate values of each point along the axes X, Y, Z, respectively, and $r_i$ is the laser reflection intensity. $t_i$ is the relative timestamp, which is optional and depends on the specific settings of a dataset. Given a set of object in the 3D scene $\mathbf{B}=\{\mathbf{b}_j=[x_j, y_j, z_j, h_j, w_j, l_j, {\theta}_j, c_j]^T \in {\mathbb{R}^{M \times 8}} \}$, where $M$ is the total number of objects, ${b}_i$ is the $i$-th object in the scene, $x_j, y_j, z_j$ is the object's center, $h_j, w_j, l_j$ is the object's size, ${\theta}_j$ is the object's heading angle and $c_j$ is the object's class. The task of 3D object detection is to detect the 3D boxes $\mathbf{B}$ from the point cloud $\mathbf{P}$ accurately.

\begin{figure*}[!htb]
  \centering
  \includegraphics[width=0.85\linewidth]{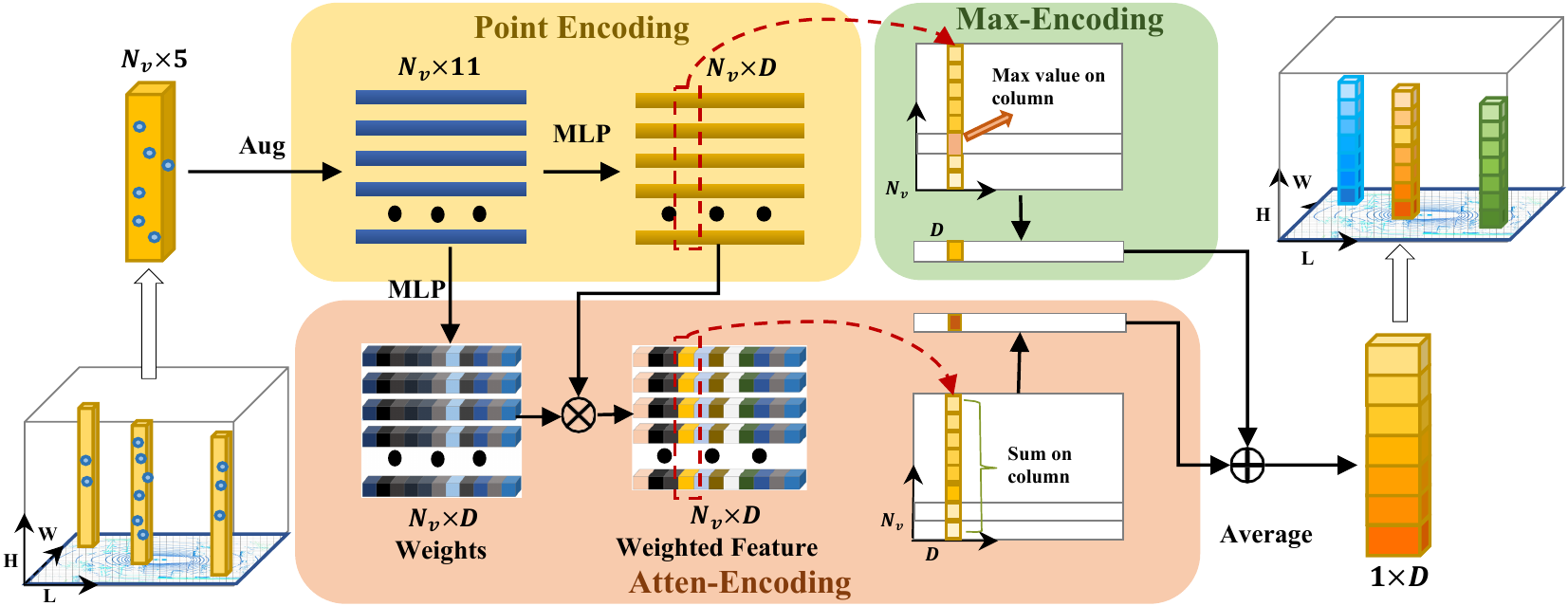}
  \vspace{-1.0mm}
  \caption{The MAPE module comprises three units: point encoding, max-pooling encoding and attention-pooling encoding. For one pillar containing $N_v$ points, in point encoding unit, we firstly augment the raw points with the pillar center and point cloud range, then map the augmented features to the feature space by a MLP.  In max-encoding unit, we obtained pillar-wise features by max-pooling operation across point dimension. In atten-encoding unit, we obtained pillar-wise features by a weighted summation operation across point dimension. The final pillar-wise feature by averaging the max- and attention-pooling features.}
 \label{fig:pillar-encoding}
 \vspace{-3.0mm}
\end{figure*}
\vspace{-1.5mm}

\vspace{-0.5mm}
\subsection{Max-and-Attention Pillar Encoding}
\label{subsec:MAPE} 

Point cloud voxel/pillar encoding is very crucial for grid-based 3D detection methods. The pioneering PointPillars aggressively utilizes max pooling to aggregate point features in each pillar. However, the max-pooling operation will result in the loss of fine-grained information, and those local geometric patterns are vital for pillar-based objects, especially for small objects. Therefore, paying attention to this information is important for accurate detection.

In this paper, we propose a simple yet efficient pillar encoding module, named \textbf{Max-and-Attention Pillar Encoding (MAPE)}, which takes into account every pillar's local detailed geometric information with negligible computational burden and benefits the performance of small objects (\textit{e.g.}, pedestrian and cyclist). Meanwhile, the lightweight MAPE module makes it highly suitable for real-time embedded applications. As shown in Fig.~\ref{fig:pillar-encoding}, our MAPE module consists of three units: 1) the point encoding, 2) the max-pooling encoding, 3) and the attention-pooling encoding.
 
We suppose that a point cloud $\mathbf{P}$ in the 3D space has the range of  $L$, $W$, $H$ along the axes X, Y, Z. $\mathbf{P}$ is equally divided into a specific pillar grid with the size of $l$, $w$, $H$. Here, as in PointPillar, we only voxelize point cloud in the XY plane without the height dimension. Let $v=\{p_i=[x_i, y_i, z_i, r_i, t_i] \in {\mathbb{R}^{N_v\times 5}} \}$ be a non-empty pillar contains $N$ points with the spatial shape $[l, w, H]$, $i \in \left \{ 1,..., N_v \right \}$, $N_v$ is number of points in pillar $v$. 

\textbf{Point Encoding.} First, we augment the points in each pillar into $\hat{p_i}=\{[x_i, y_i, z_i, r_i, t_i, {x}_{i}^{c},{y}_{i}^{c}, {z}_{i}^{c}, {x}_{i}^{r}, {y}_{i}^{r}, {z}_{i}^{r}] \in {\mathbb{R}^{N_v \times 11}} \}$, where $[x_i, y_i, z_i]$ is the original point coordinates in the ego frame, $[{x}_{i}^{c},{y}_{i}^{c}, {z}_{i}^{c}]$ is the offset of $p_i$ from the current pillar center, and $[{x}_{i}^{r}, {y}_{i}^{r}, {z}_{i}^{r}]$ is the relative coordinates of $p_i$ obtained by subtracting the range of point cloud. Notably, in each pillar, we did not adopt any sampling strategy to keep the number of points within each the same, because this operation may drop useful points and impair the original geometric patterns. Second, the augmented point-wise features $\hat{p_i}$ within $v$ are mapped to the high-dimensional feature space through an MLP layer. 
This process is formulated as
 
 \vspace{-2.0mm}

\begin{equation}
{p}_{i}^{e}=m(\hat{p_i}; w_m),
\end{equation}
where $m(\cdot)$ denotes an MLP, $w_m$ denotes learnable weights of function $m(\cdot)$, and ${p}_{i}^{e} \in {\mathbb{R}^{N_v \times D}} $ is point-wise feature.

\textbf{Max-pooling Encoding.} This unit aggregates all point features within a pillar into a single feature vector, while remaining invariant to point permutations in each pillar, which is formulated as

 \vspace{-3.0mm}
\begin{equation}
{f}^{m}=max({p}_{i}^{e}),
\end{equation}

where $max (\cdot)$ means the max-pooling operation across these point features, and ${f}^{m} \in {\mathbb{R}^D}$ is the resulting feature vector of each pillar.

\textbf{Attention-pooling Encoding.} This unit is designed to maintain the local fine-grained information. Max pooling is hard to integrate point-wise features ${p}_{i}^{e}$ within each pillar $v$ as it only takes the maximum value. However, the rich local detailed patterns are highly valuable for smaller object detection from BEV perspective. Therefore, we turn to the powerful attention mechanism to automatically learn the important local features. \textbf{First}, we use a function $g(\cdot)$ consisting of a shared MLP to predict attention scores for these points in a pillar, \textit{i.e.},  $s_i=g({p}_{i}^{e}; w_g)$,
where $w_g$ denotes learnable weights of the MLP, $s_i \in {\mathbb{R}^{N_v \times D}} $ is the attention scores. \textbf{Second,} the learnt attention scores can be regarded as a soft mask which dynamically weight the point-wise feature ${p}_{i}^{e}$. Finally, the weighted summed features are as follows:
 
\vspace{-2.5mm}
\begin{align}
{f}^{a}=\sum s_{ij}\cdot p_{ij}^{e},\ \sum_{i=1}^{N}s_{ij}=1
\end{align}
\vspace{-0.2mm}
where ${f}^{a} \in {\mathbb{R}^{N_v \times D}}$ is resulting pillar attention pooling features, $j \in \left \{ 1,..., D \right \}$ is the feature dimension index, $s_{ij}$ and $p_{ij}^{e}$ is the attention score and feature of $j$-th dimension at $i$-th point respectively. Notably, the weighting operation across different points (\textbf{$N_v$}), this process introduces interaction between different points inside a pillar (\textit{e.g.}, local area).
Finally, we combine the learnt pillar-wise max and attentive features by averaging them, \textit{i.e.},  
$f =\frac{{f}^{m} + {f}^{a}}{2}$,
where $f\in {\mathbb{R}^{1\times D}}$ is the final pillar-wise feature including the global-aware and local-aware information inside one pillar. The max-pooling operation preserves the maximum response feature in each pillar, while the attention pooling features maintain the local fine-grained information. By combining the two features, richer information can be effectively retained to enhance the pillar representation. Despite being simple, our MAPE module significantly improve the performance of small objects (+1.6 mAPH L2 for pedestrian) as shown in Tab.~\ref{tab:pooling_ablation} and Fig.~\ref{fig:MAPE}. 

\begin{figure}[!htb]
  \centering
  \includegraphics[width=0.85\linewidth]{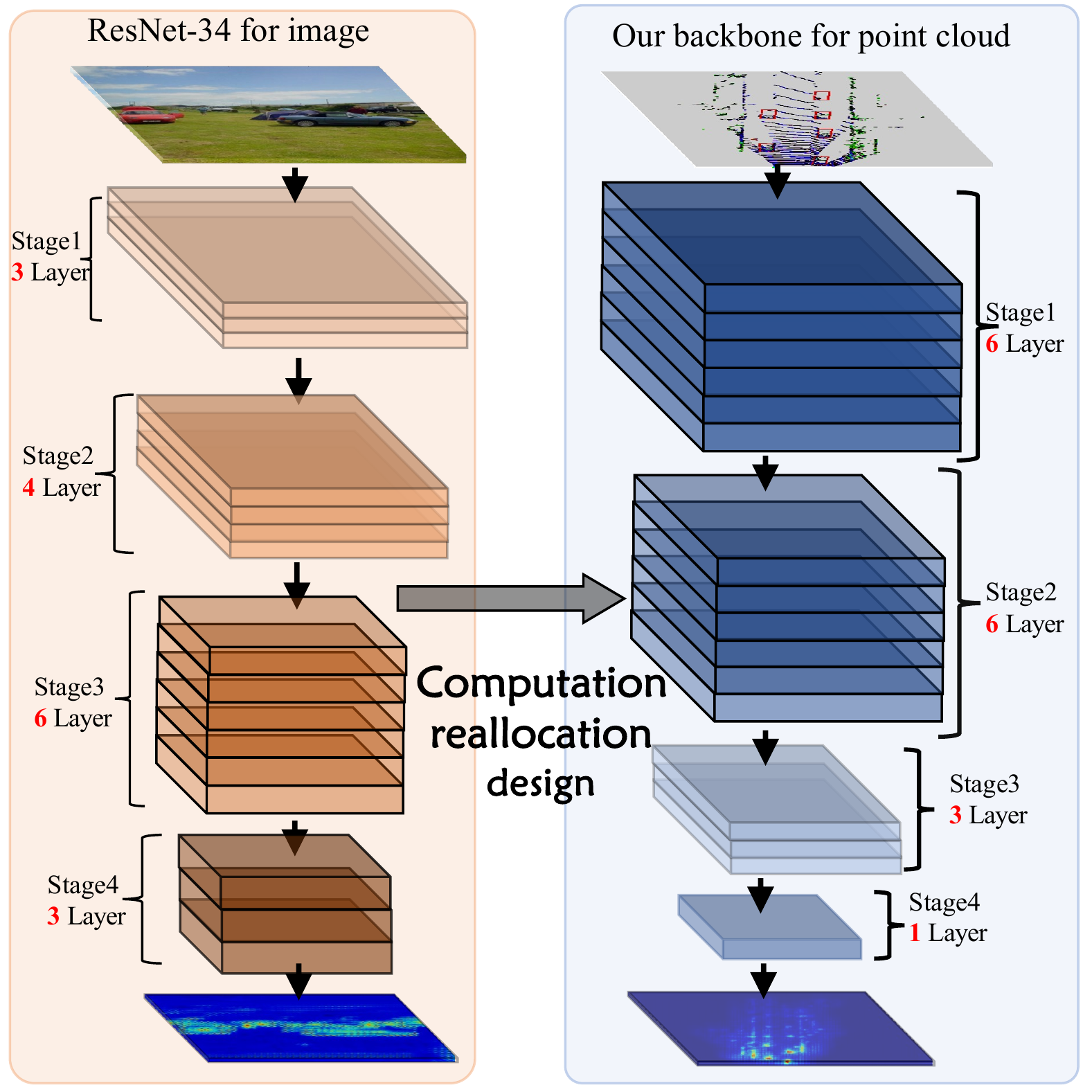}
  \caption{The diagram of our computation reallocation in backbone design. This design brings obvious performance gain without extra latency}
  \label{fig:reallocation}
  \vspace{-2.0mm}
\end{figure}
\vspace{-3.0mm}

\begin{figure}[!htp]
  \centering
  \includegraphics[width=0.95\linewidth]{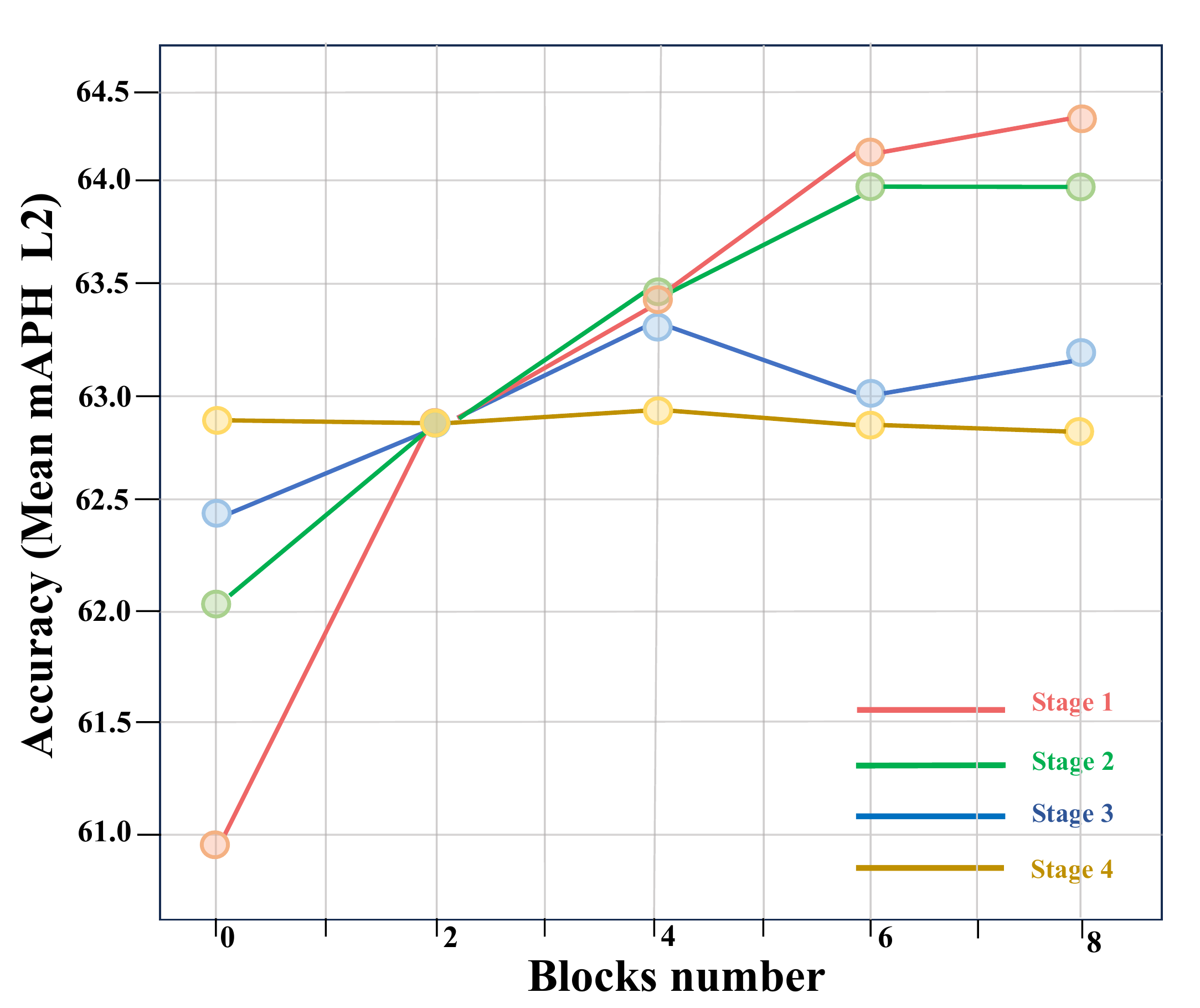}
  \caption{Performance verse different blocks number on Waymo \texttt{val} set trained with 20\% training data. Taking stage1 as an example, we arrange the number of blocks in stage1 from 0 to 8 with stride 2, while setting the number of blocks in other stages to 2.}
  \label{fig:backbone_ratio}
  \vspace{-3.0mm}
\end{figure}

\subsection{Backbone Design}
\label{subsec: backbone design}
\textbf{Computation Reallocation}. As presented in PillarNet, a lightweight and powerful backbone is very important for effective pillar feature learning. Through adopting classical 2D CNN backbones, i.e., VGGNet ~\cite{simonyan2014very}, ResNet-18/34 ~\cite{he2016deep}, PillarNet achieves a better scalability and flexibility for model complexity. However, we argue that these backbones are specifically designed to extract high-level semantic features and obtain geometries of objects in RGB image, which may not be suitable for LiDAR point cloud. This is because there exists a substantial modal difference between RGB image and LiDAR point cloud. Different from RGB image, LiDAR point cloud can easily gauges spatial distances, relationships and shapes of objects through collecting laser measurement signals to represent 3d models and maps of environments. This means that rich and accurate geometric information of objects is already explicitly encoded in LiDAR point cloud, which is also discussed in FCOS-LiDAR~\cite{tian2022fully}. Therefore, we assume that instead of allocating too much computation resources to model the geometries of objects in the later stages like ResNet in RGB images, we should reallocate the capacity to the early stages to better incorporate the geometry information carried by the raw points. 

To verify the above assumption, we systematically study computation allocation in backbone design through adjusting the stage compute ratio of ResNet. As shown in Fig.~\ref{fig:backbone_ratio}, we set ResNet-18 as the initial model, and arrange the number of blocks from 0 to 8 with stride 2 in each stage, respectively. From the results, we can find that the performance is quite sensitive to the capacity of stage (1, 2), rather than stage (3, 4), which confirms our hypothesis that it is more beneficial to allocate computation in the early stages in pillar-based LiDAR detection. Hence, to achieve a better trade-off between performance and latency, we set the stage compute ratio of four stages to (6, 6, 3, 1), which has a comparable computation cost to ResNet-34 with (3, 4, 6, 3). As shown in Fig.~\ref{fig:reallocation} and Tab.~\ref{tab:ratio}, our backbone after computation reallocation improves accuracy (+1.3 mAPH L2). See supplements for more details.

\begin{figure}[!htb]
  \centering
  \includegraphics[width=0.65\linewidth]{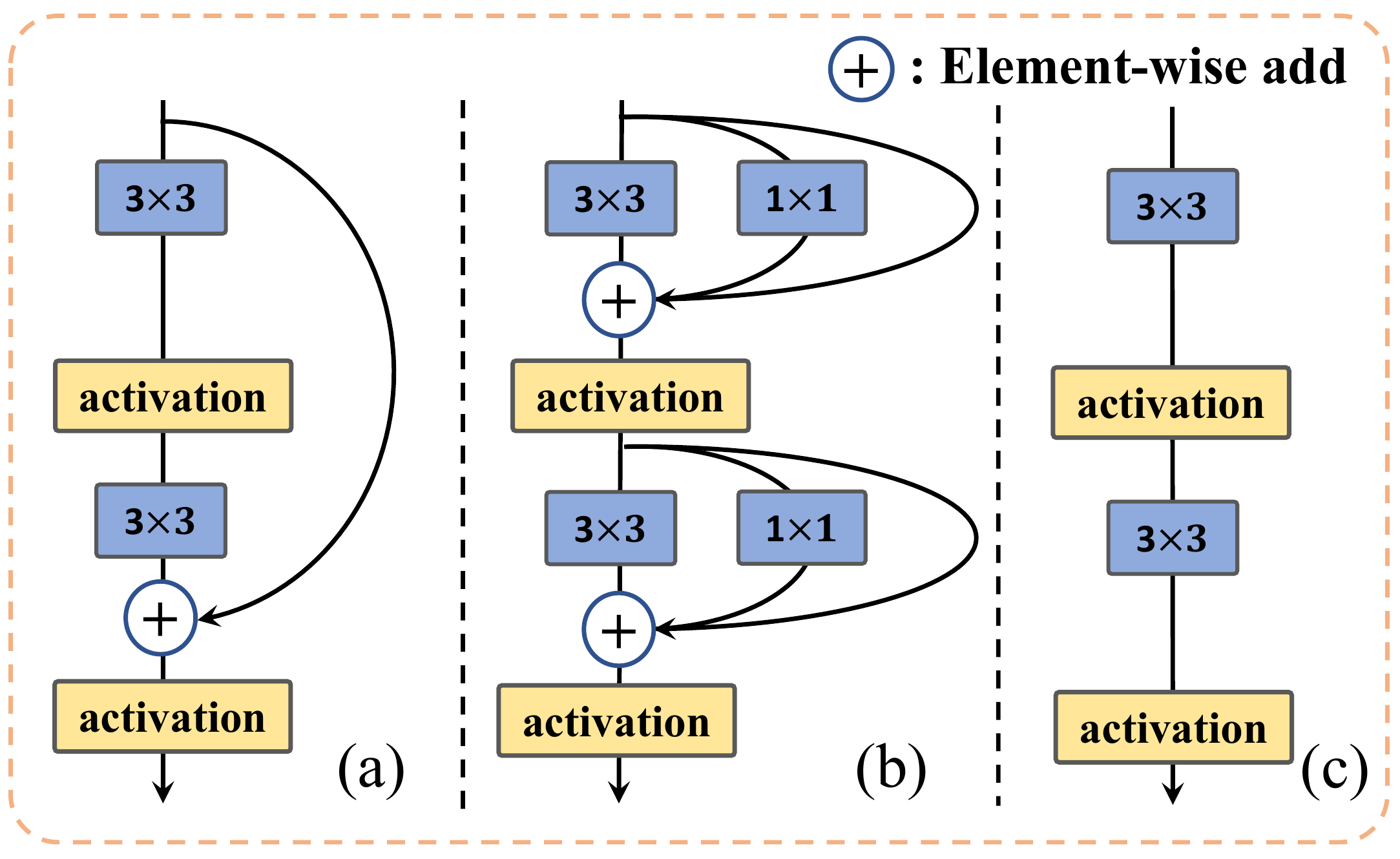}
  \caption{The design of lightweight backbone. (a) a ResNet blocks. (b) a structural re-parameterization block, which using 3 $\times$ 3, 1 $\times$ 1 and identity branches to replace a 3 $\times$ 3 conv.  (c) During inference time, a structural re-parameterization block is converted to a 3 $\times$ 3 conv.}
  \label{fig:rep}
\end{figure}

\textbf{Lightweight Backbone Design}. Re-parameterized structure~\cite{ding2021repvgg} has excellent feature representation ability in training, and can effectively reduce inference latency without performance loss through re-parameterized operation. This structure has been proven effective in 2D detection tasks~\cite{xu2022pp, li2022yolov6, wang2022yolov7}. However, they have not yet been well exploited for the 3D point cloud detection. Inspired by this, we introduce this advanced design from 2D object detection into our backbone designed for point cloud. As shown in Fig.~\ref{fig:rep}, during training, a plain res block (Fig.~\ref{fig:rep} (a)) is replaced by its over-parameterized three-branch counterpart, \textit{i.e.}, Conv3$\times3$, Conv1$\times1$ and Identity, followed by the ReLU function after the summation of the three branches (Fig.~\ref{fig:rep} (b)). The three-branch structure substantially helps the network optimization while the reparameterization converts three branches identically into one at inference (from Fig.~\ref{fig:rep} (b) to Fig.~\ref{fig:rep} (c)), improving the efficiency in inference. As a result, as shown in Tab.~\ref{tab:backbone}, our backbone network with single-path structure reduces inference latency (14\%) while improving accuracy (+0.6 mAPH/L2).  During training, a plain Conv-BN-ReLU is replaced by its over-parameterized three-branch counterpart, \textit{i.e.}, Conv3$\times3$, Conv1$\times1$ and identity, followed by the ReLU function after the summation of the three branches. 

\vspace{+1.0mm}
\textbf{The Insight of Backbone Design}. 
\vspace{-1.0mm}
\begin{itemize}
    \item Our core insight is computation reallocation backbone design for point cloud based on root modality difference of point cloud and images. Specifically, we conclude that we should reallocate the capacity to the early stages (1, 2) to better integrate the geometric information carried by the raw points, instead of allocating the capacity in the later stages (3, 4) like ResNet setting. This provides a novel perspective and principle to design backbone architecture for point clouds to the community. 
    \item For re-parameterized structure, which demonstrated that simply adapting the advanced backbone design from 2D object detection brings non-trivial improvements to 3D object detection, which is encouraging to explore more successful practices in the image domain to upgrade the network designs for point clouds.
 \end{itemize}

\begin{table*}[t]
\centering
\resizebox{\textwidth}{!}{%
\begin{tabular}{lcccccccccc}
        \toprule
        \multirow{2}{*}{Method} &\multirow{2}{*}{Reference}& \multirow{2}{*}{\#Frames}  & Latency & Speedup & Mean L2 & Vehicle L2 & Pedestrian L2 & Cyclist L2 & \#MACs & FPS\\
        & &  & (ms) & (\citeyear{yin2021center}) & (mAPH) & (mAP/APH) & (mAP/APH) & (mAP/APH) & (G)\\ 
        \midrule
         \textcolor{darkgray!50}{SECOND\textsuperscript{1}}&\textcolor{darkgray!50}{Sensors~\citeyear{yan2018second}} & \textcolor{darkgray!50}{1}  & \textcolor{darkgray!50}{--} & \textcolor{darkgray!50}{--} & \textcolor{darkgray!50}{57.2} & \textcolor{darkgray!50}{63.9 / 63.3} & \textcolor{darkgray!50}{60.7 / 51.3} & \textcolor{darkgray!50}{58.3 / 57.0} & \textcolor{darkgray!50}{--} &\textcolor{darkgray!50}{--}\\
         \textcolor{darkgray!50}{PointPillars\textsuperscript{1}}&\textcolor{darkgray!50}{CVPR~\citeyear{lang2019pointpillars}} &\textcolor{darkgray!50}{1}  & \textcolor{darkgray!50}{--} & \textcolor{darkgray!50}{--} & \textcolor{darkgray!50}{57.8} & \textcolor{darkgray!50}{63.6 / 63.1} & \textcolor{darkgray!50}{62.8 / 50.3} & \textcolor{darkgray!50}{61.9 / 59.9} &\textcolor{darkgray!50}{--} &\textcolor{darkgray!50}{--} \\
       \midrule
        CenterPoint&CVPR~\citeyear{yin2021center}& 1  & 64.3 & 1.0$\times$ & 65.5 & 66.7 / 66.2 & 68.3 / 62.6 & 68.7 / 67.6 & 307.9 &15.5\\
        PillarNeXt &CVPR~\citeyear{li2023pillarnext} & 1  & 103.2 & 0.6$\times$ & 69.1 & 70.3 / 69.8 & 74.9 / 69.8 & 70.6 / 69.6 & 281.0 &9.7 \\
        FSD &NeurIPS~\citeyear{fan2022fully} & 1  & 74.3 & 0.9$\times$ & \textbf{69.7} & 68.9 / 68.5 & 73.2 / 68.0 & 73.8 / 72.5 & - &13.5\\
        VoxelNeXt &CVPR~\citeyear{chen2023voxenext} & 1  & 48.9 & 1.3$\times$ & 68.2 & 69.7 / 69.2 & 72.2 / 65.9 & 70.7 / 69.6 & 38.7 &20.4\\
        PillarNet &ECCV~\citeyear{shi2022pillarnet} & 1  &38.7  & 1.7$\times$ &67.2 & 70.4 / 69.9 & 71.6 / 64.9 & 67.8 / 66.7 & 319.5 &25.8 \\
         \rowcolor{gray!55}
        \textbf{FastPillars} &Ours& 1  & \textbf{36.5} & \textbf{1.8$\times$} & 69.3 & 71.5 / 71.1 & 73.2 / 67.2 & 70.5 / 69.5 & 894.7 &27.4\\
        \midrule
        CenterPoint&CVPR~\citeyear{yin2021center} & 2  & 72.2 & 1.0$\times$ & 68.4 & 67.7 / 67.2 & 71.0 / 67.5 & 71.5 / 70.5 & 318.6 &13.9\\
        PillarNet&ECCV~\citeyear{shi2022pillarnet} & 2  & 45.6 & 1.6$\times$ & 70.0 & 71.6 / 71.1 & 74.5 / 71.4 & 68.3 / 67.5 & 331.0 &21.9\\
         \rowcolor{gray!55}
        \textbf{FastPillars} &Ours & 2 & \textbf{41.2} & \textbf{1.8$\times$} & \textbf{72.5} & 72.5 / 72.0 & 75.5 / 72.4 & 73.9 / 73.0 & 895.2 &24.3 \\
        \midrule
        CenterPoint&CVPR~\citeyear{yin2021center} & 3  & 80.8 & 1.0$\times$ & -- & -- & -- & --  & 325.7 &12.4\\
        CenterPoint++&CVPR~\citeyear{centerpoint++} & 3  & 50.1 & 1.6$\times$ & 71.6 & 71.8 / 71.4 & 73.5 / 70.8 & 73.7 / 72.8 & 294.6 &19.9\\
        \rowcolor{gray!55}
        \textbf{FastPillars} &Ours& 3  & \textbf{46.0} & \textbf{1.8$\times$} & \textbf{73.3} & 73.2 / 72.8 & 76.3 / 73.2 & 74.6 / 73.8 & 895.6  &21.7\\
        \bottomrule
\end{tabular}
}
\caption{Results of single-stage 3D detectors on the Waymo \texttt{val} set. FastPillars achieves 1.8$\times$ speedup over CenterPoint while being more accurate. Methods with $<$60 L2 mAPH are marked \textcolor{darkgray!50}{gray} due to the low performance.\textsuperscript{1}: from FSD paper.} 
\vspace{-2.0mm}
\label{tab:way_val}
\end{table*}

 \vspace{-2.0mm}
\subsection{Neck and Center-based Head}
 \vspace{-1.0mm}
\label{subsec: contextual instance centroid perception}
In the neck block, we followed PillarNet~\cite{shi2022pillarnet} by adopting an enriched neck design to fuse features of different levels (8$\times$ and 16$\times$) for effective interaction of spatial semantic features. In the head block, we directly utilize the center-based detection head~\cite{yin2021center}. Besides, as in AFDetV2~\cite{hu2022afdetv2}, we use an IoU branch to bridge the gap between the classification and regression prediction. 
\vspace{-1.0mm}

\subsection{Loss Functions}
\label{subsec: end-to-end learning}
We follow~\cite{yin2021center} to design our loss functions.

To be specific, for the classification branch, we use the focal loss~\cite{lin2017focal} as the heatmap loss $\mathcal{L}_{cls}$. For the 3D box regression, we make use of the L1 loss $\mathcal{L}_{reg}$ to supervise their localization offsets, size and orientation. For the IoU branch, we also utilize the L1 loss $\mathcal{L}_{iou}$ to supervise, where the target 3D IoU score $I$ is $2 \times(I - 0.5)\in [-1,1]$. Besides, the DIoU loss  $\mathcal{L}_{od-iou}$  ~\cite{zheng2020distance} is added in the regression branch. The overall loss consists of four parts as follows:
\begin{equation}
\centering
\mathcal{L}_{total} =\lambda_{1}\mathcal{L}_{cls} + \lambda_{2}\mathcal{L}_{iou} + \lambda_{3} (\mathcal{L}_{od-iou} + \mathcal{L}_{reg})
\label{eq:one-loss}
\end{equation}
where $\lambda_{1}$, $\lambda_{2}$, and $\lambda_{3}$ represent the weights of these losses.

\section{Experiments}
\label{sec:experiments}

\label{sec:dataset}
\textbf{nuScenes Dataset.} nuScenes~\cite{caesar2020nuscenes} dataset contains 700 training scenes, 150 val scenes and 150 test scenes. Each frame is generated approximately 30K points by a 32 channels LiDAR sampled with 20Hz. It contains 40K annotated key-frames and 10 categories in total. We report nuScenes detection score (NDS) and Mean Average Precision (mAP), where NDS is the main ranking metric. 

\textbf{Waymo Open Dataset.} Waymo Open Dataset~\cite{sun2020scalability} contains 1150 sequences in total, 798 for training, 202 for validation and 150 for test. Each sequence is sampled at 10Hz with a 64 channels LiDAR containing 6.1M vehicle, 2.8M pedestrian, and 67k cyclist boxes. Each frame covers a scene with a size of 150m×150m. The official evaluation tools evaluated the methods in two difficulty levels: LEVEL1 for boxes with more than five LiDAR points, and LEVEL2 for boxes with at least one LiDAR point.

\label{sec:implementation-details.}
\textbf{Implementation Details.} FastPillars use the same training schedules as prior methods~\cite{shi2022pillarnet} with the Adam optimizer under the Det3D~\cite{zhu2019class} framework on 8 A100 GPUs. We use one-cycle learning rate policy~\cite{smith2019super} with an initial learning rate 10e-4, weight decay 0.01, and momentum 0.85 to 0.95. For a fair comparison, we follow~\cite{yin2021center} to only use double-flip test-time augmentation without any model ensemble on the nuScenes Dataset. Besides, we also use the ground-truth copy-paste data augmentation from ~\cite{yan2018second} during training and disable this data augmentation in the last 5 epochs following ~\cite{wang2021pointaugmenting} (\textit{e.g.}, fade strategy). 


\begin{table*}[!thb]
\centering
\resizebox{\textwidth}{!}{%
\begin{tabular}{c|c|c|c|cc|cc|cc|cc|cc|cc}
\hline\noalign{\smallskip}
\multirow{2}{*}{Methods} &\multirow{2}{*}{Reference} & \multirow{2}{*}{Stages} & \multirow{2}{*}{Frames}  & \multicolumn{2}{c|}{Vehicle (L1)} & \multicolumn{2}{c|}{Vehicle (L2)} & 
\multicolumn{2}{c|}{Ped. (L1)} & \multicolumn{2}{c|}{Ped. (L2)} & 
\multicolumn{2}{c|}{Cyc. (L1)} & \multicolumn{2}{c}{Cyc. (L2)} \\ 
 & & &  & mAP & mAPH & mAP & mAPH & mAP & mAPH & mAP & mAPH & mAP & mAPH & mAP & mAPH \\  
\hline
PointPillars$^{\dagger}$ &CVPR~\citeyear{lang2019pointpillars} & One  & 1 & 68.60 & 68.10 & 60.50 & 60.10 & 68.00 & 55.50 & 61.40 & 50.10 & - & - & - & - \\
RCD &CoRL~\citeyear{bewley2020range} & Two  & 1 & 71.97 & 71.59 & 65.06 & 64.70 & - & - & - & - & - & - & - & - \\
CenterPoint &CVPR~\citeyear{yin2021center} & Two  & 1 & 80.20 & 79.70 & 72.20 & 71.80 & 78.30 & 72.10 & 72.20 & 66.40 & - & - & - & - \\
PV-RCNN &CVPR~\citeyear{shi2020pv} & Two  & 1 & 80.60 & 80.15 & 72.81 & 72.39 & 78.16 & 72.01 & 71.81 & 66.05 & 71.80 & 70.42 & 69.13 & 67.80 \\
AFDetV2 &AAAI~\citeyear{hu2022afdetv2} & One  & 1 & 80.49 & 80.43 & 72.98 & 72.55 & 79.76 & 74.35 & 73.71 & 68.61 & 72.43 & 71.23 & 69.84 & 68.67 \\
PV-RCNN++ &IJCV~\citeyear{shi2021pv} & Two & 1& 81.62 & 81.20 & 73.86 & 73.47 & 80.41 & 74.99 & 74.12 & 69.00 & 71.93 & 70.76 & 69.28 & 68.15 \\
PillarNet-34$^{\ddag}$ &ECCV~\citeyear{shi2022pillarnet} & One & 1& 81.91 & 81.47 & 74.43 & 74.01 & 80.22 & 73.43 & 74.16 & 67.78 & 69.00 & 67.79 & 66.48 & 65.35 \\
 \rowcolor{gray!55}
\textbf{FastPillars}&Ours& One  & 1& \textbf{82.75} & \textbf{82.31} & \textbf{75.42} & \textbf{75.01} & \textbf{81.05} & \textbf{74.84} & \textbf{75.09} & \textbf{69.22} &  \textbf{72.94} &  \textbf{71.82} &  \textbf{70.26} &  \textbf{69.23} \\
\midrule

CenterPoint &CVPR~\citeyear{yin2021center} & Two  & 2 & 81.05 & 80.59 & 73.42 & 72.99 & 80.47 & 77.28 & 74.56 & 71.52 & 74.60 & 73.68 & 72.17 & 71.28 \\
PV-RCNN &CVPR~\citeyear{shi2020pv} & Two  & 2 & 81.06 & 80.57 & 73.69 & 73.23 & 80.31 & 76.28 & 73.98 & 70.16 & - & - & - & - \\
VISTA &CVPR~\citeyear{deng2022vista} & One  & 2 & 81.70 & 81.30 & 74.40 & 74.00 & 81.40 & 78.30 & 75.50 & 72.50 & 74.90 & 73.90 & 72.50 & 71.60 \\
Pyramid R-CNN  &ICCV~\citeyear{mao2021pyramid} & Two  & 2 & 81.77 & 81.32 & 74.87 & 74.43 & - & - & - & - & - & - & - & - \\
AFDetV2 &AAAI~\citeyear{hu2022afdetv2} & One  & 2 & 81.65 & 81.22 & 74.30 & 73.89 & 81.26 & 78.05 & 75.47 & 72.41 & 76.41 & 75.37 & 74.05 & 73.04 \\
PillarNet-34$^{\ddag}$&ECCV~~\citeyear{shi2022pillarnet} & One  & 2 & 82.80 & 82.37 & 75.65 & 75.25 & 82.17 & 78.73 & 76.49 & 73.21 & 71.25 & 70.27 & 69.05 & 68.10  \\
PV-RCNN++ &IJCV~\citeyear{shi2021pv} & Two & 2& 83.74 & 83.32 & 76.31 & 75.92 & 82.60 & 79.38 & 76.63 & 73.55 & 74.44 & 73.43 & 72.06 & 71.09 \\
 \rowcolor{gray!55}
\textbf{FastPillars}&Ours& One  & 2& \textbf{83.60} & \textbf{83.17} & \textbf{76.52} & \textbf{76.12} & \textbf{82.82} & \textbf{79.45} & \textbf{77.20} & \textbf{73.90} & \textbf{76.52} & \textbf{75.45} & \textbf{74.14} & \textbf{73.11} \\
\midrule
3D-MAN  &CVPR~\citeyear{yang20213d} & Multi  & 16 & 78.71 & 78.28 & 70.37 & 69.98 & 69.97 & 65.98 & 63.98 & 60.26 & - & - & - & - \\
RSN   &CVPR~\citeyear{sun2021rsn} & Two  & 3 & 80.70 & 80.30 & 71.90 & 71.60 & 78.90 & 75.60 & 70.70 & 67.80 & - & - & - & - \\
CenterPoint++ &CVPR~\citeyear{yin2021center} & Two  & 3 & 82.78 & 82.33 & 75.47 & 75.05 & 81.07 & 78.21 & 75.13 & 72.41 & 74.40 & 73.33 & 72.04 & 71.01\\
SST&CVPR~\citeyear{fan2022embracing} & Two  & 3 & 80.99 & 80.62 & 73.08 & 73.74 & 83.05 & 79.38 & 76.65 & 73.14 & - & - & - & -\\
SWFormer\_3f &ECCV~\citeyear{swformer} & One  & 3 & 82.89 & 82.49 & 75.02 & 74.65 & 82.13 & 78.13 & 75.87 & 72.07 & - & - & - & -\\  
PillarNeXt &CVPR~\citeyear{li2023pillarnext} & One & 3& 83.28 & 82.83 & 76.18 & 75.76 & \bf84.40 & \bf81.44 & \bf78.84 & \bf75.98 & 73.77 & 72.73 & 71.56 & 70.55 \\    
 \rowcolor{gray!55}
\textbf{FastPillars} &Ours& One  & 3& \textbf{84.02} & \textbf{83.59} & \textbf{77.09} & \textbf{76.68} & 83.26 & 80.01 & 77.76 & 74.62 & \textbf{76.61} & \textbf{75.49} & \textbf{74.18} & \textbf{73.16} \\
\bottomrule
\end{tabular}
}
\caption{Single- and multi-frame LiDAR-only non-ensemble performance comparison on the Waymo \textit{test} set. $\dagger$ denotes the reported results from RSN~\cite{sun2021rsn}, $\ddag$: reproduced based on official codebase~\cite{shi2022pillarnet}.} 
\label{tab:waymo_test}
\end{table*}

\vspace{-1.0mm}
\subsection{Overall Results}
\label{sec:Overall results} 
Firstly, we compared FastPillars with the real-time 3D detectors on the Waymo \texttt{val} set. Secondly, we evaluated FastPillars on the Waymo and nuScenet \texttt{test} set. Finally, we ablated the MAPE and backbone design.


\begin{table}[!htb]
    \newcolumntype{g}{>{}c}
    \small
    \begin{center}
		\resizebox{\linewidth}{!}{
             \begin{tabular}{c|c|c|cc}
                \hline
            
                \tabincell{c}{Method}
                & Reference & Stages & NDS & mAP \\
                \hline
                \hline              
                PointPillars & CVPR~\citeyear{lang2019pointpillars} & One & 45.3 & 30.5 \\
                3DSSD & CVPR~\citeyear{yang20203dssd} & One & 56.4 & 42.6\\ 
                CenterPoint & CVPR~\citeyear{yin2021center} & Two & 65.5 & 58.0 \\
                FCOS-LiDAR& NeurIPS~\citeyear{tian2022fully}&One & 65.7 & 60.2 \\
                VMVF &CVPR \citeyear{fazlali2022versatile}&One  & 67.3 & 60.9 \\
                AFDetV2 & AAAI~\citeyear{hu2022afdetv2} & One & 68.5 & 62.4 \\
                UVTR-L& CVPR~\citeyear{li2022uvtr}&One  & 69.7 & 63.9 \\
                VISTA& CVPR~\citeyear{deng2022vista}&One  & 69.8 & 63.0 \\
                Focals Conv & CVPR~\citeyear{focalsconv-chen}&One  & 70.0 & 63.8 \\
  	        PillarNet& ECCV~\citeyear{shi2022pillarnet}&One  & 70.8 & 65.0 \\
  		VoxelNeXt & CVPR~\citeyear{chen2023voxenext}&One  & 71.4 & 66.2 \\
                \hline

  \rowcolor{gray!55}
  \textbf{FastPillars}&Ours&One  &\bf71.8 & \bf66.8 \\
  \hline
            \end{tabular}
        }
    \end{center}
    \caption{State-of-the-art comparisons for 3D detection on nuScenes $test$ set. The table is mainly sorted by nuScenes detection score (NDS) which is the official ranking metric.}
 \label{tab:nusc_test}
\end{table}

\textbf{Comparison with one-stage real-time methods.} We compare our FastPillars with state-of-the-art SPConv-based 3D detectors with different input frames setting. For fair comparison, we evaluate the whole latency on NVIDIA Tesla V100 GPU using FP16 precision and report the whole running time. We adopt SpConv v2.1.23~\cite{spconv2022} to execute the SPConv-based backbone. All modules after backbone are executed with TensorRT 8.6. All the methods are executed on the first 1,000 samples for 50 runs. Notably, FastPillars is 1.8$\times$ faster and 3.8 mAPH L2 higher than CenterPoint in single- and two-frame setting, and is faster and 4.9 mAPH L2 higher in overall, which shows a better trade-off between accuracy and latency in different input setting. Our effective model design make it available replacement for previous real-time state-of-the-art 3D detection methods CenterPoint, PillarNet.

\textbf{Evaluation on Waymo test set.} For a more comprehensive comparison, we also evaluate our FastPillars with published methods on the Waymo \texttt{test} set. As shown in Table~\ref{tab:waymo_test}, in single-frame input, FastPillars outperforms the previous one-stage and two-stage 3D detectors for the vehicle and pedestrian detection with remarkable performance gains (+1.44 mAPH L2 for the pedestrian). In two-frame input, FastPillars consistently show the superior performance compared with single-frame counterparts. Extensive experiments show FastPillars using computation reallocation and reparameterization-based structural design achieve superior performance in large scale Waymo Open Dataset. In three-frame input, our best model outperforms the two-stage SST~\cite{fan2022embracing} in the challenging pedestrian class with remarkable performance gains (+1.38 in mAPH L2 metrics). Notably, SST specially designs a single-stride transformer architecture for small object detection in 3D space. However, our method predict objects of all sizes at the same resolution feature map (8$\times$ strides). We owe such leading performance to the effective design of our MAPE module. FastPillars is 2.03 and 2.55 mAPH L2 ahead of SWFormer~\cite{swformer} which adopts transformer architecture in vehicle and pedestrian class with the same temporal information (3 frames) respectively.
PillarNeXt surpasses FastPillars in pedestrian class with 3-frames setting due to its smaller pillar size (0.075m). This leads to the fact that PillarNeXt has more computation cost and cannot run in real time. Morever, FastPillars perform superior performance than PillarNeXt in vehicle and cyclist class, owing to our efficient backbone design. 

\textbf{Evaluation on nuScenes test set.} In Table~\ref{tab:nusc_test}, we evaluate our FastPillars with other LiDAR-only non-ensemble methods on nuScenes \texttt{test} set. Both lines of results are better than previous ones. The extensive experiments verify the excellent performance of FastPillars even without SPConv. 

\vspace{-1.0mm}
\subsection{Ablation Experiments}
\label{sec:ablation}
We conduct ablation studies for the MAPE module and efficient backbone design to analyze their effect on the latency and performance on the Waymo \texttt{val} set. 

\begin{table}[!htb]
\resizebox{\linewidth}{!}{%
\begin{tabular}{c|cc|cc|cc}
\toprule
\multirow{2}{*}{Methods} & \multicolumn{2}{c|}{Vehicle (L2)} & \multicolumn{2}{c|}{Ped. (L2)} & \multicolumn{2}{c}{Cyc. (L2)} \\
  & mAP & mAPH & mAP & mAPH & mAP & mAPH \\ 
\hline
Max-pool  & 71.2 & 70.7 & 71.9 & 65.4  & 68.8 & 67.8 \\
\textbf{MAPE (Ours)} & 71.6  & 71.1  & 72.9  & 67.0 & 69.6  & 68.6  \\\hline
 \rowcolor{gray!55}
improvement &\bf{+ 0.4} &\bf{+ 0.4} &\bf{+ \textbf{1.0}}&\bf{+ \textbf{1.6}}&\bf{+ 0.8}&\bf{+ 0.8}\\ 
\hline
\end{tabular}
}
\vspace{-2.0mm}
\caption{MAPE ablation on Waymo $val$ set. MAPE shows remarkable improvements especially for small objects.}
 \label{tab:pooling_ablation}
\vspace{-1.0mm}
\end{table}

\begin{figure}[htb]
  \centering
  \includegraphics[width=0.9\linewidth]{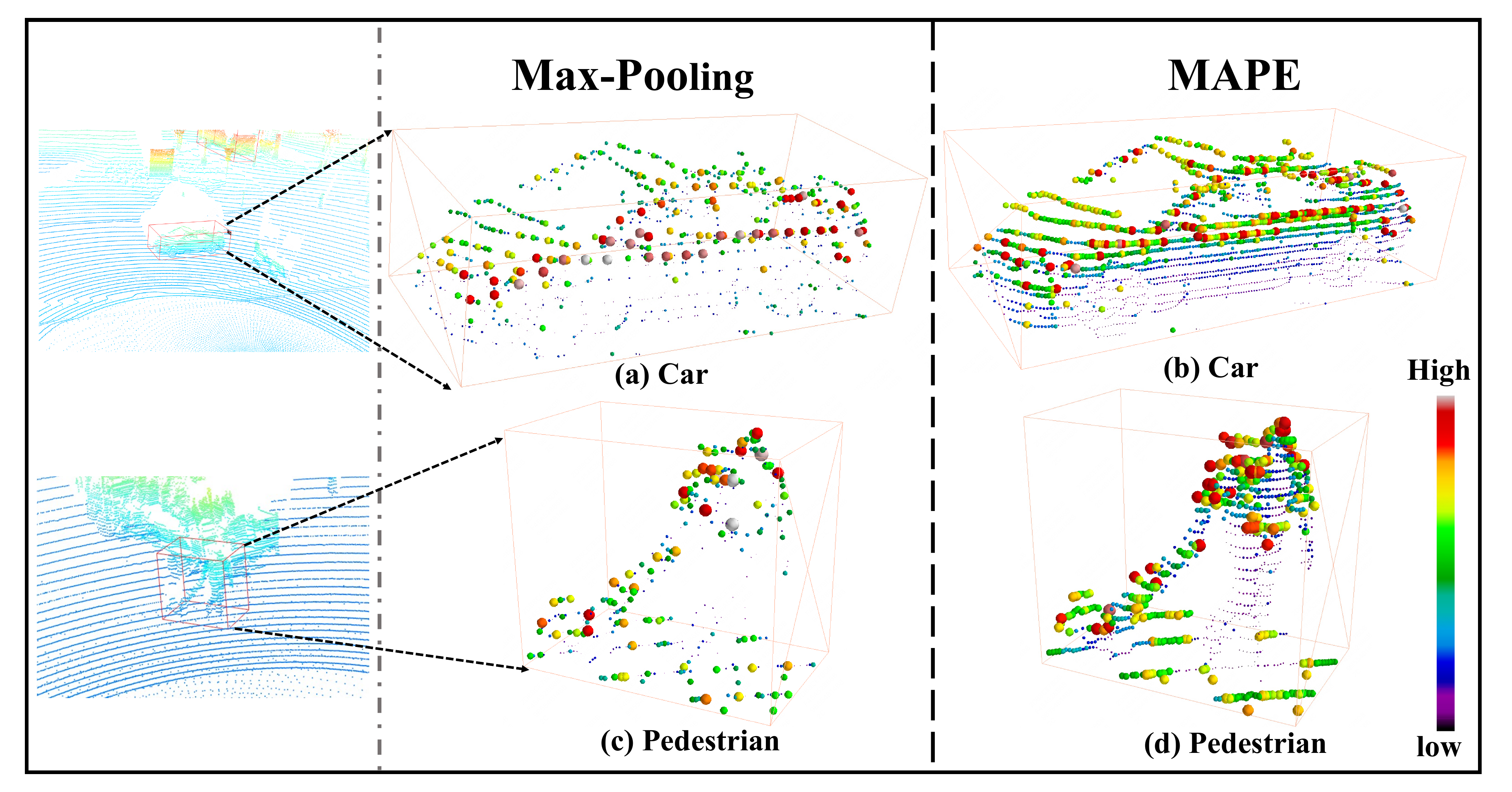}
\caption{Visualization of the attention scores on car (a-b) and pedestrian(c-d) using Max-pooling or MAPE. The size of points means scores. The point will be paid more attention if it has a higher score. Compared with (a) and (b), (c) and (d), MAPE module pays more attention to the object semantic and local geometries.  Best viewed in color.} 
  \label{fig:MAPE}
 \vspace{-3.0mm}
\end{figure}

\textbf{MAPE module}. As shown in Tab.~\ref{tab:pooling_ablation}, compared with common max-pooling operations, the MAPE module improves the performance of vehicle, pedestrian and cyclist in different degrees on Waymo \texttt{val} set. Notably, MAPE boost pedestrian category with remarkable performance gains (+1.6 mAPH L2). We further visualize the attention scores in different categories with MAPE module. As shown in Fig.~\ref{fig:MAPE}, MAPE pays more attention to the object semantic information (contour of car/pedestrian). Conversely, Max-pooling lose much geometries. The qualitative and quantitative results show that our MAPE module encodes the local fine-grained geometrical patterns of objects and the most prominent features (\textit{i.e.}, the maximum value) effectively by combining the atten- and max-pooling operation and improve the perceptual ability in the BEV perspective. More ablation studies refer to supplements.

\label{sec:pillar-encoding}
\vspace{-1.0mm}

\textbf{Computation Reallocation in Backbone.} We conduct experiments by arranging the number of blocks per stages from 0 to 8 with a stride of 2, while setting the number of blocks in other stages to 2, as shown in Tab.~\ref{tab:ratio}. We observed the different phenomena in different stages.

\begin{table}[!htb]
\huge
\centering
\resizebox{\linewidth}{!}{
\begin{tabular}{c|cc|ccc|c}
\toprule
\multirow{2}{*}{Stage} &\multirow{2}{*}{Ratio}  & {Mean (L2)} & {Veh. (L2)} & {Ped. (L2)} & {Cyc. (L2)} &{GFLOPS} \\
 & & mAPH &mAPH &mAPH &mAPH &\\ 
 \hline
\multirow{4}{*}{1} &0, 2, 2, 2  &61.0 & 67.1 & 56.1  & 59.6   &278.0\\
&2, 2, 2, 2$^\dagger$ &62.8 & 67.7  & 58.5  & 62.3  &355.1\\
&4, 2, 2, 2 & 63.4 &68.2 & 59.5  & 62.6    &432.2\\
&6, 2, 2, 2 & 64.1(+\textbf{3.1})  &68.4 & 59.9  & 64.0    &509.3\\
\hline
\multirow{4}{*}{2} &2, 0, 2, 2  &62.0 & 67.2  & 57.6 & 61.2& 278.0\\
&2, 2, 2, 2$^\dagger$ &62.8 & 67.7  & 58.5  & 62.3  &355.1\\\
&2, 4, 2, 2 &63.4 & 68.2  & 59.5  & 62.7  &432.2\\
&2, 6, 2, 2 &64.0(+\bf{2.0}) & 68.8  & 59.7 & 63.5  &509.3\\
\hline
\multirow{4}{*}{3} &2, 2, 0, 2  &62.4 & 67.0  & 58.3 & 62.0  &278.0\\
&2, 2, 2, 2$^\dagger$ &62.8 & 67.7  & 58.5  & 62.3  &355.1\\
&2, 2, 4, 2 &63.4 & 68.4  & 58.9 & 62.7  &432.2\\
&2, 2, 6, 2 &63.0(+\bf{0.6}) & 68.4  & 58.7 & 62.0 &509.3\\
\hline
\multirow{4}{*}{4} &2, 2, 2, 0  &62.9 & 67.9  & 58.9 & 61.8 &278.0\\
&2, 2, 2, 2$^\dagger$ &62.8 & 67.7  & 58.5  & 62.3  &355.1\\
&2, 2, 2, 4 &62.9 & 67.8  & 58.7  & 62.2 &432.2\\
&2, 2, 2, 6 &62.8(-\bf{0.1}) & 67.6  & 58.7 & 62.2 &509.3\\
\hline
Res-34 &3, 4, 6, 3 & 63.5 & 68.8  & 59.6  & 62.2  &660.3\\
\hline
\rowcolor{gray!55}
Ours &6, 6, 3, 1 & 64.8(+\bf{1.3})  & 69.1  & 60.8  & 64.5  &660.3\\
\hline
\end{tabular}
}
\caption{Performance of various stage compute ratio on Waymo $val$ set trained with 20\% training data. $\bf{\dagger}$ means the stage compute ratio of Res-18.}
\label{tab:ratio}
\vspace{-2.0mm}
\end{table}

\begin{itemize}
\item In stage 1, increasing the number of blocks remarkably improves the overall performance, from 61.0  to 64.1, with 3.1 mAPH L2 gain. Specifically, mAPH L2 of vehicles, pedestrian and cyclist can be improved 1.3, 3.8 and 4.4, respectively.  
\item In stage 2, increasing the number of blocks also brings a significantly improvement in the overall performance, from 62.0 to 64.0 mAPH L2. Although the improvement is relatively small compared to stage 1 (+2.0 vs. +3.1), it still shows consistent gains for different types of objects (vehicles +1.6, pedestrian +2.1, cyclist +2.3 mAPH L2). 

\item In stage 3, increasing the number of blocks brings a slight improvement in the overall performance, from 62.4 to 63.0 mAPH L2. The type of vehicle still shows a continuous improvement, from 67.0 to 68.4, while the pedestrian and cyclist have a relatively weak gain.

\item In stage 4, increasing the number of blocks yields little overall performance improvement and performance for each category is almost unchanged.

\end{itemize}
Based on the above results, we conclude that \textbf{(1)} Increasing the number of blocks in stage (1, 2, 3) can improve the overall detection performance, especially in stage (1, 2). \textbf{(2)} For different object class, the vehicle exhibits consistent improvement with the increase in the number of blocks in stage (1, 2, 3) and no improvement in stage 4. \textbf{(3)} The pedestrian and cyclist show consistent improvement with the increase in the number of blocks in stage (1, 2) and very little even negative gain in stage (3, 4). 

The results prove our previous assumption: for pillar-based LiDAR detection tasks, we should reallocate the capacity to the early stages to better integrate the geometric information carried by the raw points, instead of allocating the capacity in the later stages like ResNet setting. In particular, allocating more computing sources in the early stages can effectively improve small objects' accuracy. Therefore, to achieve a better performance under the constraint of low computation resource, we set the stage compute ratio of four stages to (6, 6, 3, 1), which has a comparable computation cost to ResNet-34 with (3, 4, 6, 3). Besides, we think that the proposed compute ratios of different stage is coarse. Considering factors such as resolution, channel dimensions, and stage depths of the backbone, neural architecture search (NAS) technique can be used to achieve improved capacity reallocation. This remains an open problem for future research to the community.

\begin{table}[!htb]
\huge
\resizebox{\linewidth}{!}{%
\begin{tabular}{c|ccccc}
\toprule
setting &Ratio & mAPH L2  \textbf{$\uparrow$} & Latency(ms)$\downarrow$ & Params(M)$\downarrow$ & FLOPs(G)$\downarrow$   \\
\hline
Vanish & 3,4,6,3& 63.5 &17.7& 24.2 & 660.3  \\  
\hline
Backbone & 6,6,3,1 & 64.8   & 17.6  & 12.1    &660.3      \\ 
+Rep  & 6,6,3,1 & 65.4  &16.0 & 12.1  & 660.3  \\ 
 \hline
  \rowcolor{gray!55}
   improve   &  & \textbf{{+ 1.9}}& \textbf{{-1.7}} &  \textbf{{ -12.1}} & \textbf{{--}}    \\ 
  \hline
\end{tabular}
}
 \vspace{-2.0mm}
\caption{Ablation on the lightweight backbone design. Our backbone can effectively improve performance (+1.9 mAPH L2) while reducing latency (11\%) and memory cost (50\%).} 
 \label{tab:backbone} 
\vspace{-3.0mm}
\end{table}

\textbf{Lightweight Backbone Design}. Tab.~\ref{tab:backbone} demonstrates the efficacy of our backbone in reducing the number of parameters (50\%) and improving performance (1.3 mAPH L2) through computation reallocation. Furthermore, we introduce the re-parameterized structure, which brings 11\% reduction in latency and a 0.6 mAPH L2 gain. Our backbone achieves remarkable 1.9 mAPH L2 improvement and 11\% acceleration, proving its effectiveness. 

\vspace{-2.0mm}
\section{Conclusion}
\label{sec:conclusion}
In this paper, we propose \textbf{FastPillars}, a real-time one-stage pillar-based 3D detector, to simultaneously improve the detection accuracy and runtime efficiency while keeping the deployment in mind. In particular, we show that SPConv can be safely sidestepped with a redesigned lightweight and effective backbone. Moreover, we also propose a Max-and-Attention Pillar Encoding (MAPE) module to compensate for the information loss in the pillar encoding.

Extensive experiments show that our FastPillars achieves a better trade-off between speed and accuracy, and can be deployed through TensorRT for real-time on-device applications. Given its effectiveness and efficiency, we hope that our method can serve as a strong and simple alternative to current mainstream SPConv-based 3D detectors. Besides, we believe that improved computation reallocation, taking into account factors such as resolution, channel dimensions, and stage depths for the backbone, could be achieved using the neural architecture search (NAS) technique. This remains an open problem for future research.
\bibliography{aaai24}

\clearpage
\appendix

\section{More Ablation Study}

\begin{table}[!htb]
\huge
\resizebox{\linewidth}{!}{%
\begin{tabular}{c|c|cc|cc|cc}
\toprule
\multirow{2}{*}{Methods} & Mean (L2)& \multicolumn{2}{c|}{Vehicle (L2)} & \multicolumn{2}{c|}{Ped. (L2)} & \multicolumn{2}{c}{Cyc. (L2)} \\
& mAPH & mAP & mAPH & mAP & mAPH & mAP & mAPH \\ 
\hline
CenterPoint-PP & 55.9& 62.5  & 62.0  & 61.8  & 51.5 & 56.1 & 54.3  \\\hline
CenterPoint-PP+  \bf{MAPE} & 57.0& 63.0  & 62.5  & 63.5 & 53.5 & 56.7 & 54.9  \\\hline
\rowcolor{gray!55}
improvement &\bf{+1.1} &\bf{+0.5} & \bf{+0.5} & \bf{+1.7}& \textbf{+2.0}&\bf{+0.6}&\bf{+0.6}\\ 
\hline
\end{tabular}
}
\caption{Apply MAPE module to CenterPoint-Pillar on Waymo $val$ set trained with 20\% training data.} 
 \label{tab:mape_ablation}
\end{table}
\vspace{-3.0mm}
\paragraph{MAPE with PointPillars.} To evaluate the generalization of the proposed MAPE module, we further apply  the proposed MAPE module to pillar encoding of CenterPoint-Pillar. As shown in Table \ref{tab:mape_ablation}, MAPE shows remarkable improvements in pedestrian class (+2.0 mAPH L2) and an overall performance gain of 1.1 mAPH L2. This result shows that by introducing attention-pooling operation in pillar encoding, the fine-grained geometrical patterns of objects can be better encoded, thus improving the performance of small objects.

\begin{table}[!hb]
\tiny
\resizebox{\linewidth}{!}{%
\begin{tabular}{cccc|cc}
\hline
& Atten & Avg & Max & mAP & NDS \\  
\hline 
(a)&\checkmark & & &  60.5 & 67.1\\
(b) & & \checkmark & &  60.5 & 66.7\\ 
(c) & & & \checkmark &  61.2 & 67.7 \\ 
(d) & \checkmark & \checkmark & &  61.2 & 67.7 \\ 
(e) &  & \checkmark & \checkmark & 59.8 & 67.0 \\
(f) & \checkmark &\checkmark   & \checkmark &  61.1 & 67.7 \\
\hline
\rowcolor{gray!55}
\textbf{MAPE} & \checkmark &  & \checkmark &  \textbf{61.5} & \textbf{68.1} \\ 
\hline
\end{tabular}%
}

\caption{Performance comparison of different pooling operations on nuScenes $val$ set.}
\label{tab:Ablation-MAPE}
\end{table}

\vspace{-3.0mm}
\paragraph{Different Pooling Operations in MAPE.} We compare the performance of different pooling operations in Tab.~\ref{tab:Ablation-MAPE}. Only using attention-pooling operation achieves 60.5 map performance, while only using max-pooling operation has higher 61.2 map. The reason may be that the max-pooling operation retains the most representative feature, while the attention-pooling focuses on the fine-grained information of different points. However, the combination of  Atten- and Max-pooling (\textit{i.e.}, MAPE) obtains the best performance with a mere 4ms of added latency compared with commonly used max-pooling operation in PointPillars~\cite{lang2019pointpillars}. This is because that our proposed MAPE module can not only effectively learn the most representative features, but also pay attention on local geometrical patterns automatically, which is beneficial to small objects.

 \section{Detailed Breakup Time of FastPillars}
 
\begin{table}[!htb]
\resizebox{\linewidth}{!}{
\begin{tabular}{cccccccc}
\hline
Device &Method    & PFE/VFE  & Backbone &Neck-Head & P-P &Overall &FPS\\ \hline
 \multirow{4}{*}{V100
 -32G}&CenterPoint-1f   &20.7 & 29.5 &7.9 & 6.2  &64.3 &15.5\\

&\textbf{FastPillars-1f}    &6.7 & 16.0 & 8.4 & 5.4  &36.5 & \textbf{27.4}\\ 
\cmidrule(lr){2-8} 
&CenterPoint-2f   &25.2 & 32.3 &7.9 & 6.8  &72.2 &13.9\\
&\textbf{FastPillars-2f}    &10.9 & 16.0 & 8.4 & 5.9  &41.2 & \textbf{24.3}\\ \hline
 \multirow{2}{*}{A100
 -80G} &\textbf{FastPillars-1f} &5.1 & 14.3 & 7.0 & 4.5 & 30.9  &\textbf{32.4}\\  
 &\textbf{FastPillars-2f} & 9.2 & 14.3 &7.0 & 5.1 & 35.3  &\textbf{28.1}\\ \hline
\end{tabular}%
}
\caption{Inference time on Waymo $val$ set.} 
 \label{tab:real-time compare}
\end{table}

As shown in Tab.~\ref{tab:real-time compare}. We evaluated the average running time on Waymo val set to measure the detailed inference time of FastPillars in different devices. For example, FastPillars-1f achieved 32.4 FPS on a single NVIDIA A100 GPU, including 5.1 ms for pillar encoding, 21.3 ms for model forward propagation, and 4.5 ms for post-processing.

\vspace{-1.0mm}
 \section{FastPillars Deployment on Edge Devices}
  \vspace{-1.0mm}
\begin{figure}[htb]
  \centering
  \includegraphics[width=\linewidth]{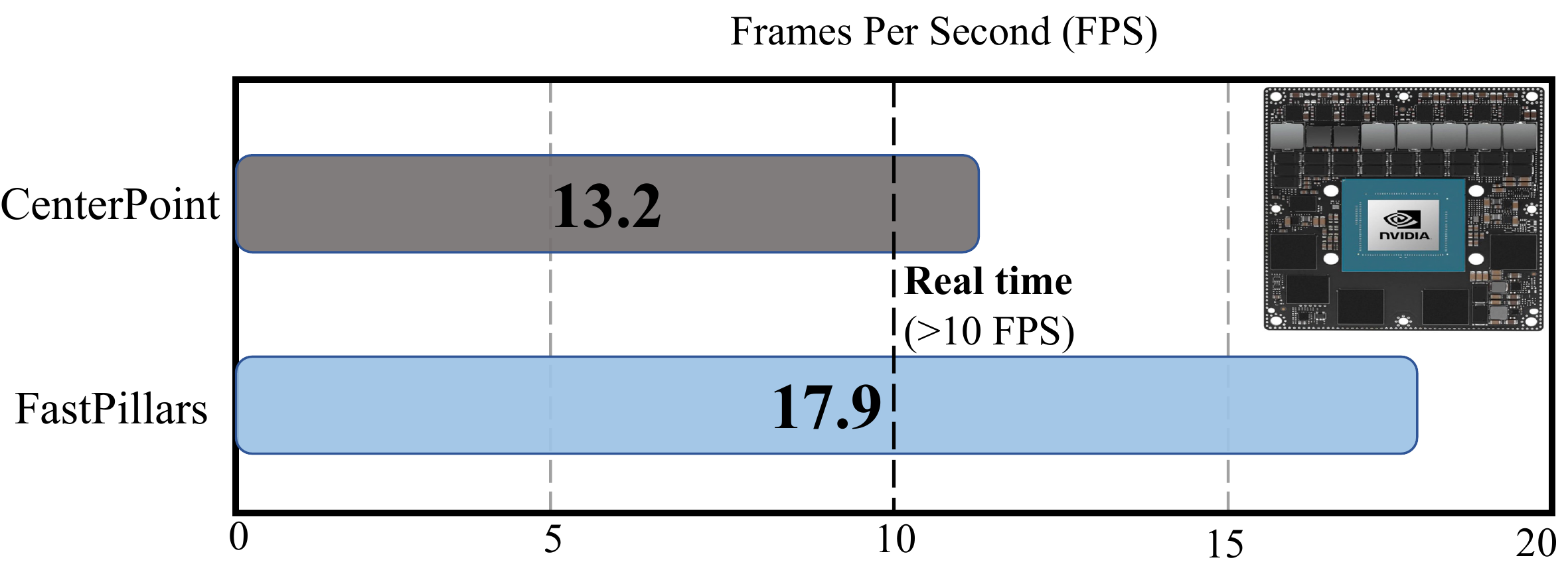}
\caption{Measured latency on NVIDIA Jetson AGX Orin. FastPillars can run in real-time on edge GPUs.} 
 \label{fig:Orin}
  \vspace{-1.0mm}
\end{figure} 

SPConv is not a built-in operation in TensorRT. This makes it necessary to write a tedious custom plugin in CUDA C++ with several limitations like fixed-shape input and reduced compatibility for commonly-used TensorRT for the quantization deployment. Therefore, the use of SPConv makes it hard to be quantized and deployed via TensorRT. In contrast, our model can be easily exported as the standard ONNX/TRT format, allowing it to run on edge devices where TensorRT is supported. Furthermore, we deploy our FastPillars on an NVIDIA Jeston AGX Orin, a resource-constrained edge GPU platform widely used in real-word autonomous driving. As show in Fig.~\ref{fig:Orin}, FastPillars runs at 18 FPS, is faster than CenterPoint (both report the network forward time). This shows that FastPillars can be deployed on different types of edge hardware (\textit{i.e.}, the \textbf{deployment-friendly ability}). We believe that it serves as a strong and simple alternative to current mainstream SPConv-based 3D detectors for efﬁcient LiDAR-centric perception in real-world deployment applications.

\begin{figure*}[!htb]
  \centering
  \includegraphics[width=0.95\linewidth]{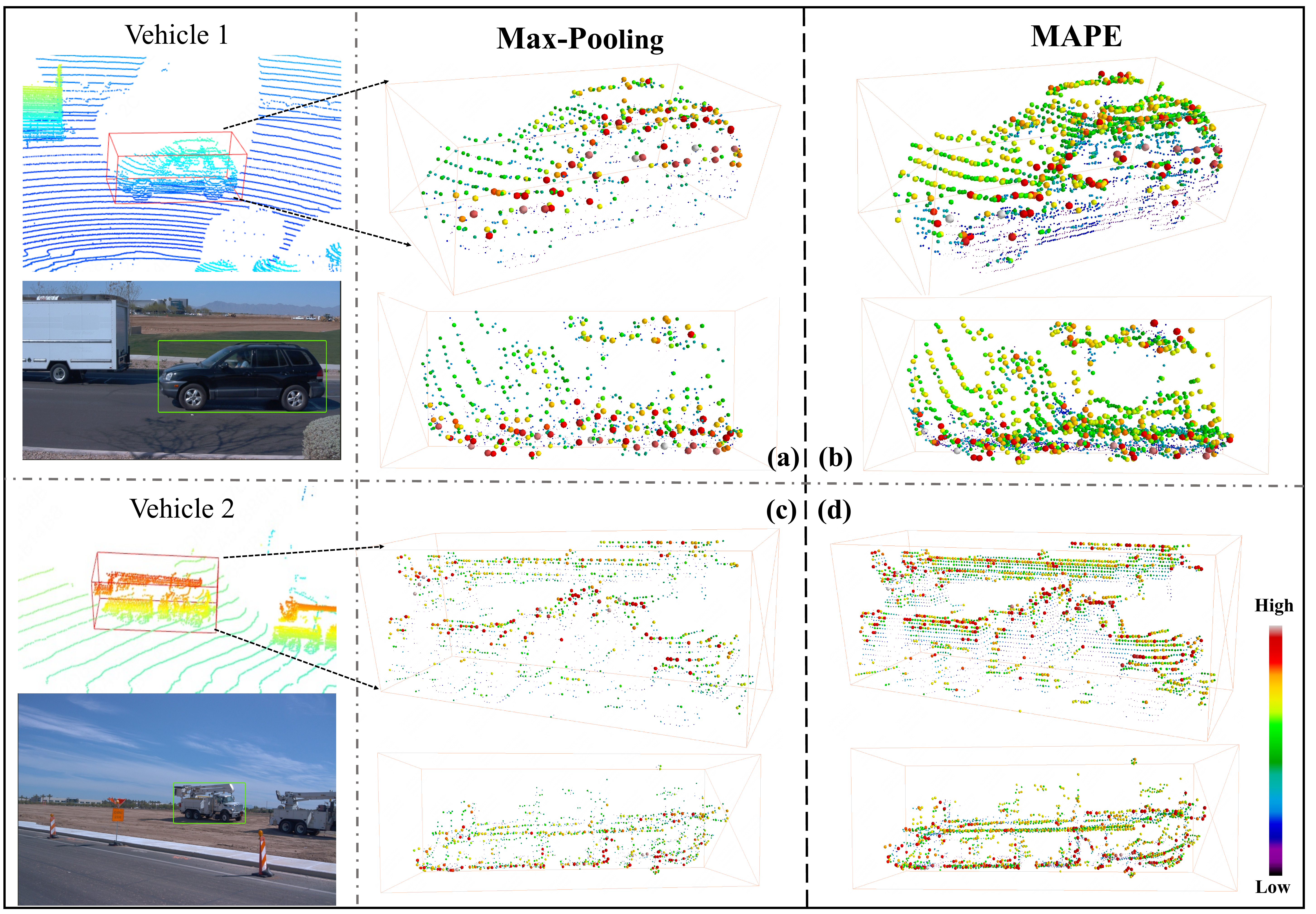}
   \includegraphics[width=0.95\linewidth]{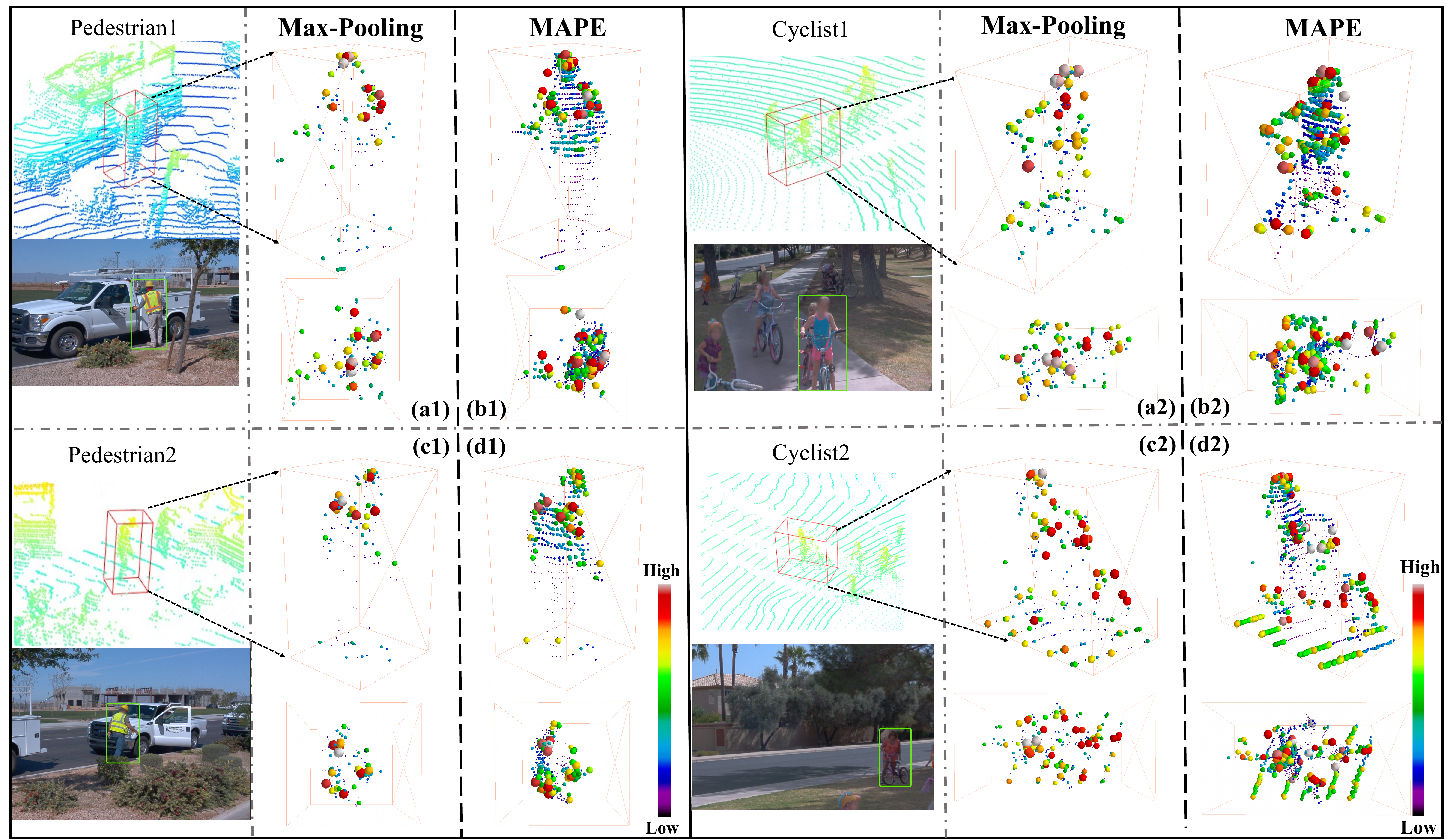}
\caption{Visualization of the learned attention scores in MAPE module on different class. The size and color of points represents their scores. The point will be paid more attention if it has a higher score. Best viewed in color.}
  \label{fig:MAPE_veh}
\end{figure*}

\vspace{-2.0mm}
\section{More Implementation details}
 \vspace{-2.0mm}
We use one-cycle learning rate policy~\cite{smith2019super} with an initial learning rate 10e-4 during training. The learning rate gradually increases to 0.001 in the first 40\% epochs and then gradually decreases to 10e-5 in the rest of the training process. The whole point cloud is flipped randomly along the X or Y axis, randomly rotated along the Z axis in the range $[-\pi/4, \pi/4]$ and translated by $[-0.5m, 0.5m]$, as well as globally scaled by a random factor sampled from $[0.95, 1.05]$. In loss function, $\lambda_{1}$, $\lambda_{2}$, $\lambda_{3}$ are set to 1.0, 1.0 and 0.25, respectively. For backbone, the number of channels in the four stages is 64, 128, 256, 512.

For nuScenes dataset, we set the detection range of the point cloud to $[-54m, 54m]$ for the X and Y axis and $[-5m, 3m]$ for the Z axis and set the pillar size as 0.15m. Besides, following CenterPoint we use the class-agnostic NMS with the score threshold set to 0.2 and rectification factor $\alpha$ to 0.5 for 10 classes during the post-processing in inference. FastPillars is trained by 20 epochs, which takes $\sim$25 hours on 8 A100 GPUs with batch size 32.

For Waymo Open Dataset, we set the detection range of the point cloud to $[-75.2m, 75.2m]$ for the X and Y axis and $[-2m, 4m]$ for the Z axis and set the pillar size as 0.2m. We train the FastPillar from scratch with batch size 32, max learning rate 3e-3 for 36 epochs. For the post-processing process during inference, following AFDetV2~\cite{hu2022afdetv2}, we use class-specific NMS with the IoU threshold set to 0.8, 0.55, 0.55 and rectification factor $\beta$ to 0.68, 0.71, 0.65 for Vehicle, Pedestrian and Cyclist respectively.
\vspace{-1.0mm}

\section{More Qualitative Analysis of MAPE}
\vspace{-1.0mm} 
For \textbf{vehicles} class, we visualize two cars in different views (point-view and bird eye view) in Fig.~\ref{fig:MAPE_veh}. Compared with (a) (b) and (c) (d), the MAPE module obviously pays more attention to the object semantic information (car's outline) and local geometric context information. However, the Max-pooling operation loses much useful fine-grained information. Notably, FastPillars is a BEV-based method rather than a point-based method. For \textbf{pedestrian} and \textbf{cyclist} class, due to the tiny size and non-rigid property, pedestrian and cyclist detection are more challenging than vehicle detection. Pedestrian and car detection are quite different. In Fig.~\ref{fig:MAPE_veh}, we visualize pedestrian and cyclist categories in different views (point-view and bird eye view). Compared with (a1) (b1) and (c1) (d1), the MAPE module focus more stable geometric feature in pedestrian (pedestrian's body parts), and has richer representation than the Max-pooling under BEV. Similar with pedestrian, Compared with (a2) (b2) and (c2) (d2), cyclist also maintain more semantic information with the MAPE modules. Different from the Max-pooling operation, MAPE reserves some background information, which enhances the local context representation for detection. Experiments demonstrate the superiority and effectiveness of our MAPE module, which introduces low latency and is suitable for real-time deployment applications.

\vspace{-3.0mm} 
\begin{figure*}[!htb]
 \centering
 \includegraphics[width=\linewidth]{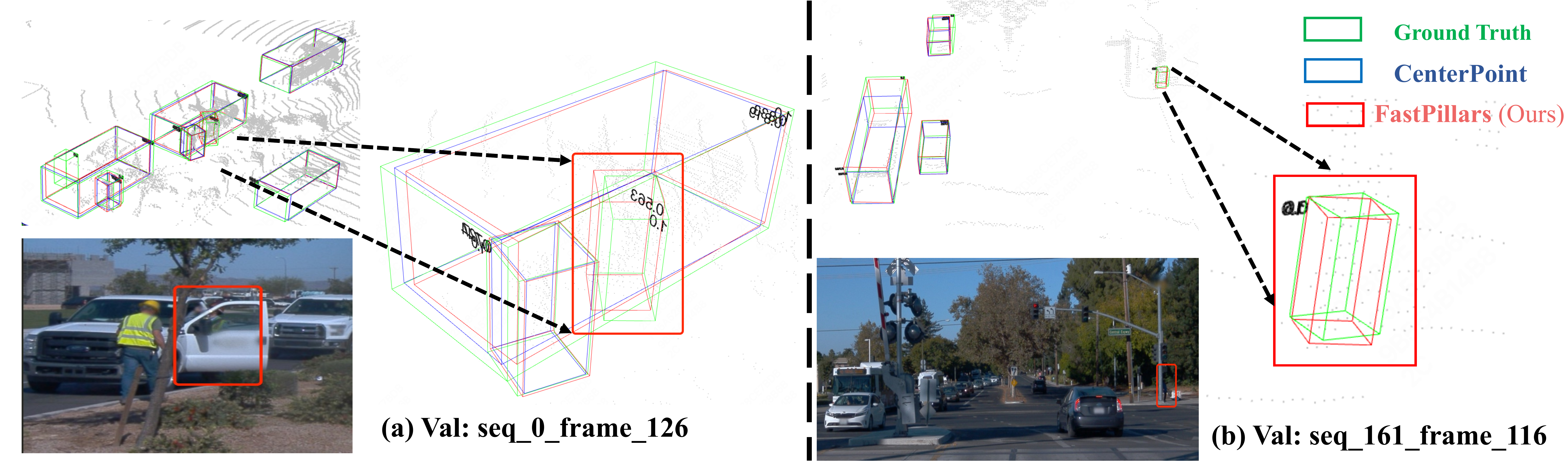}
 \caption{Qualitative Experiments of FastPillars on the Waymo $val$ set with baseline (CenterPoint~\cite{yin2021center}) . \textcolor{blue}{Blue} boxes mean CenterPoint~\cite{yin2021center} predictions,  \textcolor{green}{Green} boxes and \textcolor{red}{red} boxes are ground-truth and FastPillars predictions, respectively. Best viewed in color.}
 \label{fig:vis_centerpoint_waymo}
\end{figure*}

\begin{figure*}[!htb]
  \centering
  \includegraphics[width=\linewidth]{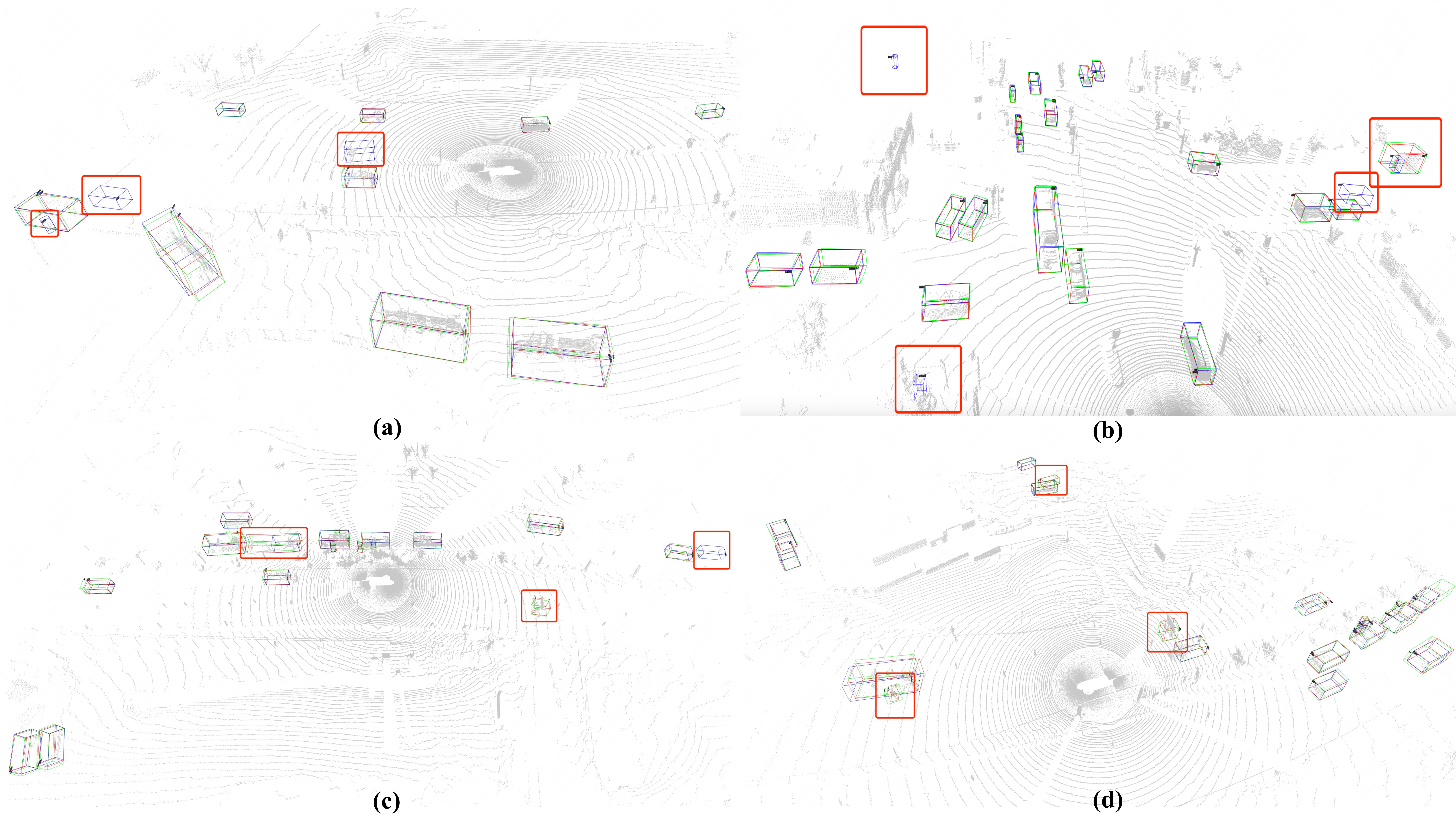}
  \caption{Qualitative Experiments of FastPillars on the Waymo $val$ set. \textcolor{blue}{Blue} boxes mean Pillarnet~\cite{shi2022pillarnet} predictions,  \textcolor{green}{Green} boxes and \textcolor{red}{red} boxes are ground-truth and FastPillars predictions, respectively. Best viewed in color.}
 \label{fig:vis_quantative_waymo}
\end{figure*}

\begin{figure*}[htb]
  \centering
  \includegraphics[width=\linewidth]{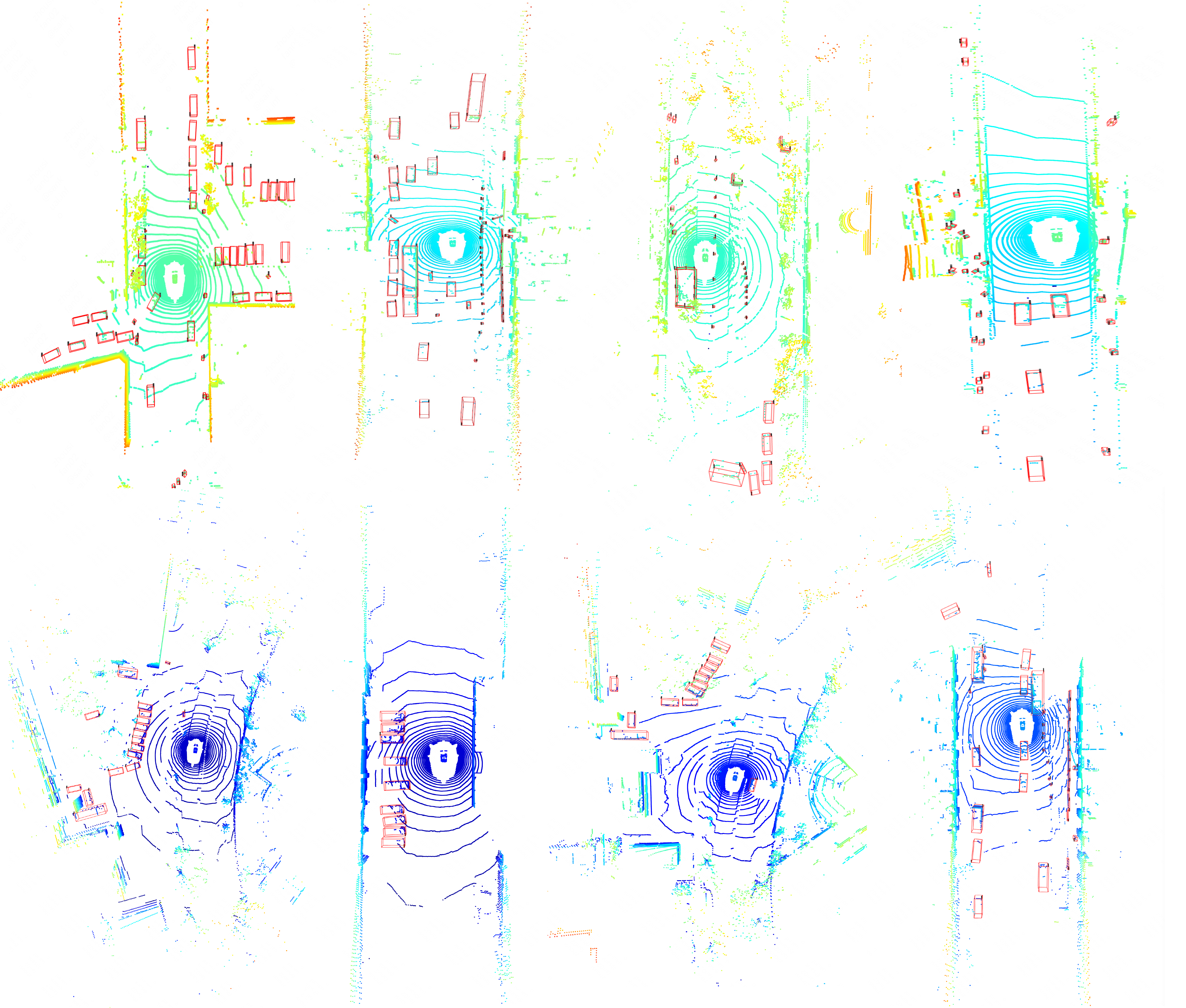}
  \caption{Visualization results of FastPillars on the nuScenes $val$ set.}
 \label{fig:vis_nus}
\end{figure*}
 
 \begin{figure*}[htb]
  \centering
  \includegraphics[width=\linewidth]{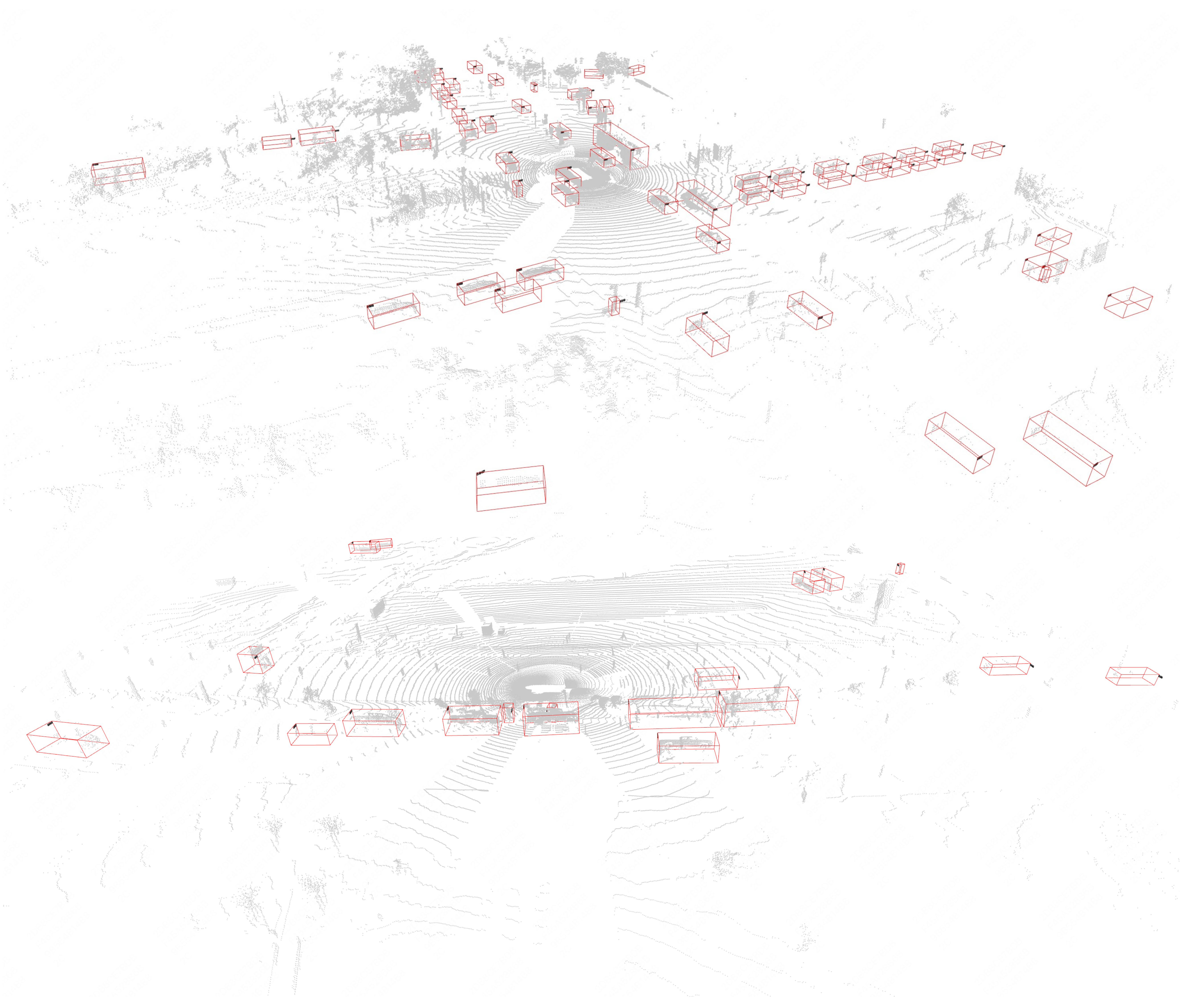}
  \caption{Visualization results of FastPillars on the Waymo $val$ set.}
 \label{fig:vis_waymo1}
\end{figure*}

\begin{figure*}[htb]
  \centering
  \includegraphics[width=\linewidth]{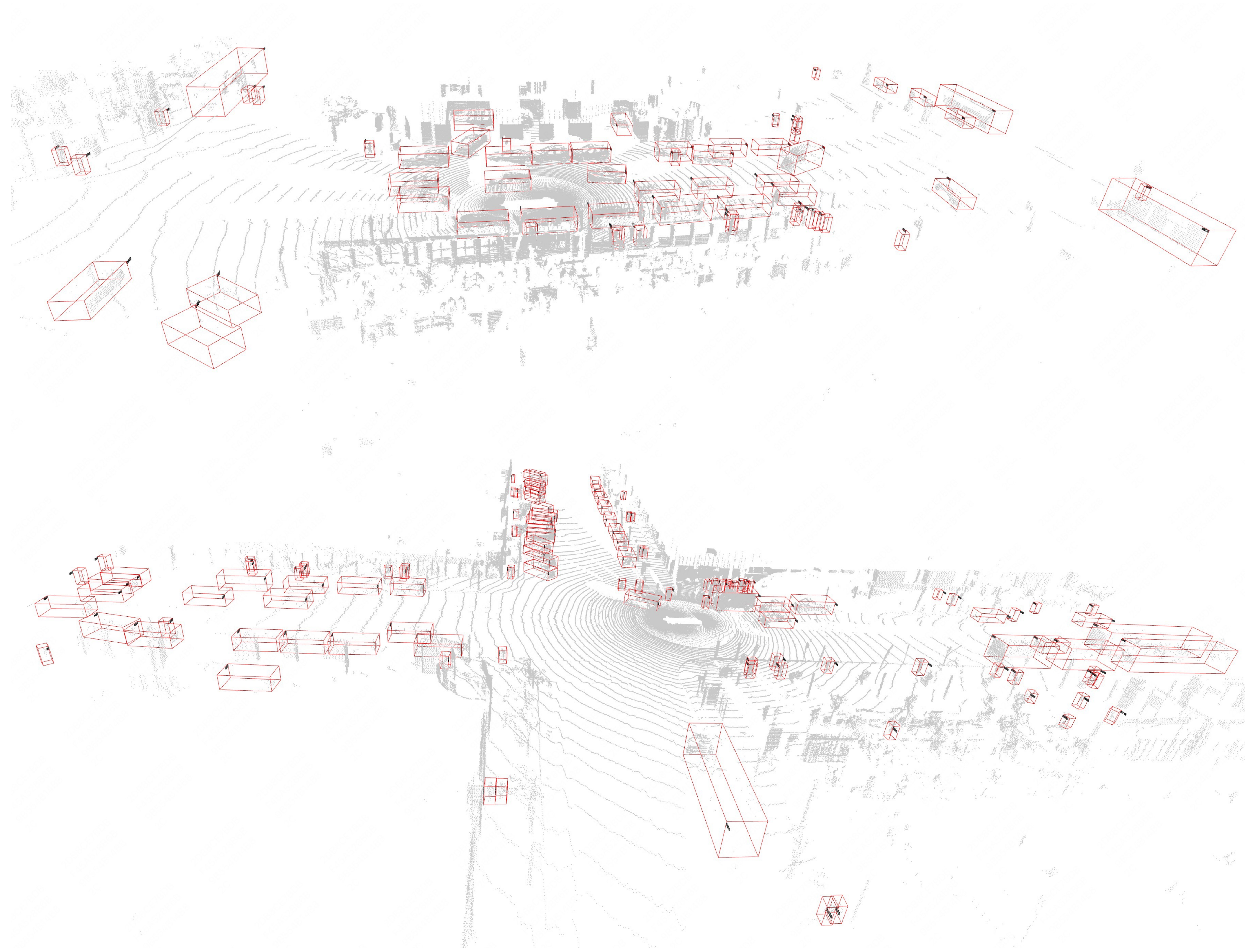}
  \caption{Visualization results of FastPillars on the Waymo $val$ set.}
 \label{fig:vis_waymo2}
\end{figure*}
 
 \section{Visualization Results with Baseline}
 
As shown in Fig.~\ref{fig:vis_centerpoint_waymo}, we visualize results on the Waymo \texttt{val} set, demonstrating that our method outperform baseline (CenterPoint), especially in pedestrian classes. This shows the effectiveness of our MAPE and backbone design.

\section{Qualitative Experiments on Waymo Dataset} %
Here, we make a qualitative comparison with previous state-of-the-art SPConv-based method PillarNet on Waymo $val$ set. As shown in Fig.~\ref{fig:vis_quantative_waymo}, PillarNet has a lot of false detection and missed detection, but FastPillars has better detection performance.

 \vspace{-1.0mm}
\section{Visualization on nuScenes Dataset}
\vspace{-1.0mm}

Some visualization results are shown in Fig.~\ref{fig:vis_nus}. Here, we visualize the detection results in some challenging scenarios on the nuScenes $val$ set based on our FastPillars. As we can see, FastPillars can work reliably under a wide variety of challenging circumstances.  We can clearly see that the proposed FastPillars is capable of detecting small targets, such as pedestrians, barriers and bicycles.
 
\vspace{-1.0mm}
\section{Visualization on Waymo Dataset}
\vspace{-1.0mm}
We visualize the detection results on Waymo $val$  Set in Fig.~\ref{fig:vis_waymo1} and Fig.~\ref{fig:vis_waymo2} based on FastPillars model. Thanks to the powerful pillar feature encoding ability of CRVNet and fine-grained geometric information provided by the MAPE module, our FastPillars performs well on the large scenes and can locate 3D objects with sparse points accurately.

\vspace{-1.0mm}
\section{Potential Negative Social Impact}
\vspace{-1.0mm}
In this paper, we proposed a real-time pillar-based detector capable of achieving promising low-latency objects detection in autonomous driving scenarios. Our model is trained and evaluated totally based on public datasets, and there is no known potential negative impact on society.

\end{document}